\documentclass[conference]{IEEEtran}
\usepackage{cite}
\usepackage{amsmath,amssymb,amsfonts}
\usepackage{algorithmic}
\usepackage{graphicx}
\usepackage{bm}
\usepackage{textcomp}
\usepackage{xcolor}
\definecolor{customblue}{RGB}{0, 102, 204}
\usepackage{multirow}
\usepackage{makecell}
\usepackage{float} 
\usepackage{booktabs}\usepackage{graphicx}
\usepackage{subcaption}
\usepackage{array}
\usepackage[
    colorlinks=true,
    linkcolor=black,     
    citecolor=black,     
    filecolor=black,
    urlcolor=black
]{hyperref}
\def\BibTeX{{\rm B\kern-.05em{\sc i\kern-.025em b}\kern-.08em
    T\kern-.1667em\lower.7ex\hbox{E}\kern-.125emX}}
\begin{document}
\pagestyle{plain}
\title{CrashSplat: 2D to 3D Vehicle Damage Segmentation in Gaussian Splatting}

\author{
    Drago\c{s}-Andrei Chileban, 
    Andrei-\c{S}tefan Bulzan, 
    Cosmin Cern\v{a}zanu-Gl\v{a}van\\
    %\textit{Department of Computer and Information Technology} \\
    \textit{Politehnica University of Timi\c{s}oara} \\
    {\footnotesize\texttt{dragos-andrei.chileban@student.upt.ro, stefan.bulzan@student.upt.ro, cosmin.cernazanu@cs.upt.ro}}\\
}
% 3\textsuperscript{rd}
\maketitle
\thispagestyle{plain}

\begin{abstract}
\par
Automatic car damage detection has been a topic of significant interest for the auto insurance industry as it promises faster, accurate, and cost-effective damage assessments. However, few works have gone beyond 2D image analysis to leverage 3D reconstruction methods, which have the potential to provide a more comprehensive and geometrically accurate representation of the damage. Moreover, recent methods employing 3D representations for novel view synthesis, particularly 3D Gaussian Splatting (3D-GS), have demonstrated the ability to generate accurate and coherent 3D reconstructions from a limited number of views.
\par
In this work we introduce an automatic car damage detection pipeline that performs 3D damage segmentation by up-lifting 2D masks. Additionally, we propose a simple yet effective learning-free approach for single-view 3D-GS segmentation. Specifically, Gaussians are projected onto the image plane using camera parameters obtained via Structure from Motion (SfM). They are then filtered through an algorithm that utilizes Z-buffering along with a normal distribution model of depth and opacities.
\par
Through experiments we found that this method is particularly effective for challenging scenarios like car damage detection, where target objects (e.g., scratches, small dents) may only be clearly visible in a single view, making multi-view consistency approaches impractical or impossible. The code is publicly available at: \url{https://github.com/DragosChileban/CrashSplat}.
\end{abstract}


\section{Introduction}
\par Vehicle damage detection involves using image analysis to automatically detect damage to car body components. Recent advances in deep neural networks have led to solutions for this task based on supervised learning, which require large-scale labeled datasets. Some studies proposed manually-annotated datasets~\cite{WangXinkuang2023, HuynhNhan2023} while others rely on synthetically generated samples \cite{ParslovJens2024}. Most of the existing methods experiment with different detection and segmentation networks in order to compare training results on publicly available damage datasets. However, by focusing solely on 2D image analysis, they hamper the vehicle visualization and inspection process, hence restricting the overall understanding of the damage produced. In order to overcome this limitation, we propose a solution that segments damaged parts in 2D and projects the masks onto a 3D reconstruction of the vehicle. 

\begin{figure}[t]
\centering
\includegraphics[width=0.9\linewidth]{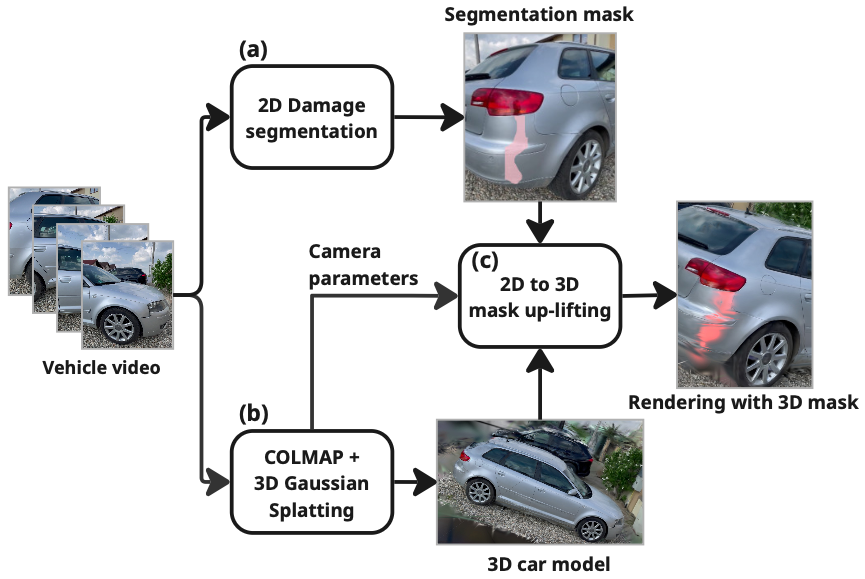}
\caption{\label{fig:gauss} Pipeline diagram. (a) Instance segmentation module that generates 2D damage masks from input frames. (b) View-synthesis generation module that computes SfM camera parameters and builds a 3D Gaussian Splatting of the vehicle. (c) 3D-GS segmentation module that projects the 2D mask onto the 3D reconstruction.}
\label{fig:diagram}
\end{figure}

\par For generating the reconstruction we use 3D Gaussian Splatting (3D-GS)~\cite{KerblBernhard2023}, a novel-view synthesis method that allows us to build a 3D car model from a limited number of images in just a few minutes. Consequently, our goal is to find the 3D Gaussians that correspond to the segmented damage masks. Current 3D-GS segmentation solutions introduce learnable Gaussian parameters that are optimized during training, requiring additional processing time~\cite{CenJiazhong2025,CenJiazhong2025_2,ChoiSeokhun2024,ZhouShijie2024,ZhuRunsong2025}. Some other studies propose learning-free approaches that do not require scene-specific training~\cite{GuoYansong2025, HuXu2025}. Using segmentation foundation models like Segment Anything Model (SAM)~\cite{KirillovAlex2023} they generate multiple masks that are used for global alignment and multi-view consistency. However such an approach would not be reliable for some use-cases (e.g. car damage detection) where consistent masks would be hard to predict across multiple views. Thus, we propose a single-view segmentation method inspired by the rasterization algorithm of 3D-GS. The Gaussians are projected on the image plane and traversed in ascending order of depth while cumulating their weights in a grid-like buffer, in the same manner as Z-buffering. This way, foreground gaussians will progressively cover the ones in the background. Remaining outliers will be filtered out using a statistical filtering of depths and opacities.
\par Our complete pipeline is described in Fig.~\ref{fig:diagram}. We first predict 2D damage masks using an instance segmentation model based on YOLO~\cite{RedmonJoseph2016} architecture. Then, we compute camera parameters using Structure from Motion (SfM)~\cite{SchonbergerJohannes2016} in order to generate a 3D Gaussian Splatting of the scene. Finally, we use the camera parameters to project the Gaussians onto the images and select only the ones that correspond to the detected mask.

\par
In summary, our work offers the following two main contributions:
\par\textbullet{ We propose an end-to-end pipeline for automatic damage detection by treating this task from a 3D perspective. Our approach improves the damage evaluation process by combining the accurate visualization offered by images with the superior geometrical representation of the damage in 3D reconstructions. We believe that the various applications of vehicle damage detection could benefit from our solution.}
\par\textbullet{ We introduce a low-complexity single-view 3D Gaussian Splatting segmentation method for up-lifting 2D masks to 3D. Extensive experiments demonstrate the robustness of our method on car damage detection and similar tasks, where multi-view segmentation masks are difficult to obtain. Our approach is particularly effective in these challenging scenarios, generating segmentation masks that successfully capture the target object and are consistent across multiple views.}

\section{Related work}

\subsection{Car automatic damage detection}

\par Automatic damage detection is strongly related to the existence of high-quality datasets for car damage classification, detection and segmentation. To our knowledge, there are a few publicly available datasets that provide detection and segmentation labels for vehicle damage. CarDD~\cite{WangXinkuang2023} is a dataset that contains 4000 images with over 9000 instances of damaged parts. The labels cover 6 damage classes that are well-balanced in the number of samples and pixel size. VehiDE~\cite{HuynhNhan2023} provides over 14000 pictures with damaged car parts from 8 categories. Another related dataset is TartesiaDS~\cite{perez2024automated}, which contains over 100 images with labels for scratches and deformations, designed for insurance company use cases. The larger of these datasets, CarDD and VehiDE, have been used by many automatic damage detection studies, being an important contribution to the topic. Recently, CrashCar101~\cite{ParslovJens2024} proposed a procedural damage generation pipeline by introducing augmented damage into 3D car models, thus obtaining synthetic 2D images with labels for car parts and damage categories. This synthetic approach can overcome the inefficiencies of manual annotation, enabling the creation of large-scale damage datasets. However, it may fail to accurately reproduce vehicle damage due to the discrepancy between real and synthetic generated data.

\par The improvement of deep neural networks has revolutionized the field of computer vision, enabling machines to interpret visual data with remarkable accuracy. Over the past decade, advances in architectures such as Convolutional Neural Networks and, more recently, transformer-based models have significantly improved performance in tasks like image classification, object detection, and semantic segmentation, paving the way for reliable automatic damage detection systems in real-world applications.
\par Dwivedi et al.~\cite{DwivediMahavir2021} used a web-scraped dataset of 2000 samples for both damage classification and detection, employing YOLO alongside various encoder architectures. This approach demonstrated the ability to identify both the class and location of the damage efficiently. Furthermore, Chen et al.~\cite{ChenQiqiang2020} applied MaskR-CNN~\cite{HeKaiming2017} for road damage detection and instance segmentation, emphasizing its ability to produce accurate per-pixel masks of damage. Chen H.~\cite{ChenHanxiao} uses Mask R-CNN for detecting damaged regions and image alignment in order to identify the modifications produced by the incident. Lee D.~\cite{LeeDonggeun2024} is introducing a three quarter view dataset with annotations for vehicle orientation and damage type, comparing results from multiple backbone networks. Experiments using YOLO architecture are performed by~\cite{Perez-ZarateSergio2024,RamazhanMuhammad2025,YogeshwaranS2025} on existing or newly introduced damaged car parts datasets, using various data augmentation techniques. 
The vast majority of these mentioned studies provide 2D detection solutions by training multiple detection architectures on existing datasets and comparing their results. In contrast, our work goes beyond 2D analysis by up-lifting segmentation masks to 3D car reconstructions, providing a more comprehensive visualization of the damage.

\subsection{3D vehicle reconstruction}
DreamCar~\cite{DuXiaobiao2024} reconstructs a complete 3D model using a limited number 
of frames (one to five). They use a 3D-GS approach and prior-knowledge of car shapes from a generative model. Their method is useful for autonomous driving scenarios, where accurate representation of 3D data is needed for bridging the gap between the 2D and 3D world and using the existing 2D large scale datasets.
Car-GS~\cite{LiCongcong2025} is using 3D-GS as well, but this time focusing on eliminating the reflecting and transparent surface artifacts by introducing other useful learnable
parameters to each Gaussian. This work can help in scenarios where high quality representations are necessary, thereby mitigating the effects of imperfect, noisy, or low-quality visual input data. Another 3D car reconstruction approach is introduced by~\cite{AuclairAdrien2007} that uses Structure-from-Motion on video frames capturing vehicles in translation. Wang et al.~\cite{WangBo2024} use a 3D wireframe library and car structure prior knowledge in order to build a 3D car model from a single view. The solution introduced by~\cite{JayawardenaSrimal} uses 3D pose estimation in order to identify damaged parts by comparing 3D CAD models of damaged and undamaged vehicles. Similar to our method,~\cite{vanRuitenbeekR} projects detected damage in 2D and projects the result on 3D car reconstructions. In comparison, they project 2D bounding boxes using ray-tracing on precomputed 3D car meshes. Car meshes are time consuming to compute and even impossible to recreate from a limited number of vehicle images. Additionally, we project segmentation masks instead of bounding boxes, which allow us to perform the projection at pixel-level accuracy.

\subsection{3D segmentation in Gaussian Splatting}
\par The task of creating multi-view synthesis from a limited number of pictures of an object is quite recent and has become a central problem in 3D computer vision and graphics. This challenge involves generating novel views of a scene by relying on learned representations that infer geometry, appearance, and lighting. Recent breakthroughs such as Neural Radiance Fields (NeRF)~\cite{MildenhallBen2021}, 3D Gaussian Splatting~\cite{KerblBernhard2023}, and other neural rendering approaches have dramatically improved the quality and realism of synthesized views. These methods leverage deep learning to model scenes as continuous volumetric functions or compact point-based structures, enabling high-fidelity reconstructions with minimal supervision. 

\par
Segmenting 3D objects in this kind of representations is a difficult task due to the lack of available 3D datasets that offer ground truth segmentation masks. Thus, the evaluation metrics are, in general, based on comparing the 2D back-projection of the 3D mask with the ground truth 2D mask.
\par
Recently, most of the works on 3D-GS segmentation are using training-based methods. Here, the semantics of the scene are learned by using newly introduced Gaussian parameters, supervised by 2D masks that are generated by 2D foundation models (SAM~\cite{KirillovAlex2023}). These methods~\cite{CenJiazhong2025,CenJiazhong2025_2,ChoiSeokhun2024,ZhouShijie2024,ZhuRunsong2025} develop a good understanding of the scene semantics and could perform multi-object real-time segmentation, but require learnable parameters that take additional time to train. GaussianCut~\cite{JainUmangi2024} builds a graph that corresponds to the 3D scene and uses a graph-cut algorithm to do interactive segmentation. FlashSplat~\cite{ShenQiuhong2024} formulates 2D to 3D segmentation as a linear programming problem and solves it in closed form. Other methods~\cite{GuoYansong2025, HuXu2025} start from already trained 3D Splats and assign some labels (IDs) to each Gaussian. Using multi-view consistency scores or by training neural networks, they later query these labels to perform interactive 3D segmentation. This approach requires additional processing time and also accurate 2D segmentation masks from multiple view, which are not always available nor necessary (some objects require less viewing angles to be segmented than others). Our method does not require any additional global pretraining or postprocessing of the scene. We use a single-view 2D mask that we up-lift to 3D by projection and find the corresponding 3D Gaussians using a Z-buffering algorithm and statistical filtering. This way, a robust 3D segmentation, consistent across multiple views, is generated from a single view.

\section{Method}

\subsection{Preliminaries}

\begin{figure*}[ht]
\centering
\includegraphics[width=0.9\linewidth]{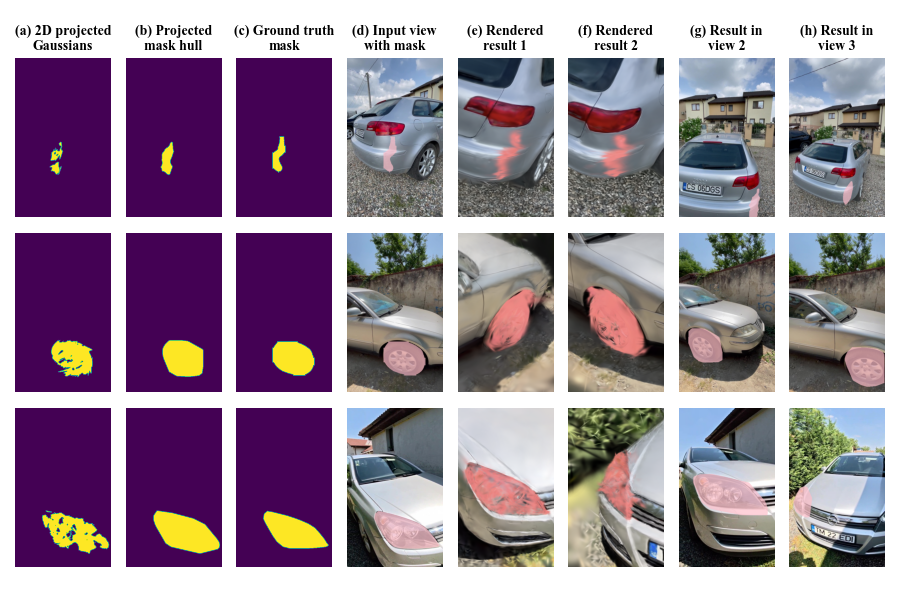}
\caption{Visual comparison of the 2D back-projected Gaussian segmentation and the ground truth mask. (a) The 3D to 2D projection of the segmented Gaussians. (b) The polygon resulted from the Convex/Concave Hull of (a). (c) The ground truth mask. (d) The view used for segmentation with the ground truth mask. (e) and (f) Rendering result of the segmentation by coloring the Gaussians. (g) and (h) The segmentation result projected in other views. Scenarios shown (top to bottom): row 1: a right-rear car scratch, row 2: a front-right wheel, row 3: a front-right car headlight.}
\label{fig:iou}
\end{figure*}

\par For the \textbf{vehicle damage instance segmentation} we use a pretrained version of the YOLO11 segmentation network from Ultralytics~\cite{JocherGlenn2023}. The model is trained on CarDD~\cite{WangXinkuang2023} and VehiDE~\cite{HuynhNhan2023} datasets that contain samples with annotated masks for diverse damage categories. We run the inference on multiple vehicle frames and generate 2D segmentation masks that will be later used by the projection algorithm. More detailed explanations and a comparison for the instance segmentation module are also included in Sec.~\ref{sec:41}.

\par \textbf{The view-synthesis generation} is achieved using multiple images capturing the vehicle from different angles. We are first using these images to compute the camera parameters by running COLMAP, which is a general-purpose Structure-from-Motion (SfM) and Multi-View Stereo (MVS) pipeline~\cite{SchonbergerJohannes2016, SchonbergerJohannes2016_2}. Here, features are extracted and matched in multiple images in order to compute the camera intrinsic and extrinsic parameters alongside a sparse point-cloud. Intrinsic parameters define how the camera maps 3D points in its local coordinate system to 2D image coordinates and are made of the focal length $(f_x, f_y)$ meaning how strongly the camera lens converges the light, and the optical center of the image $(c_x, c_y)$ (considered here to be the center of the image). These two form the intrinsic matrix:
\begin{equation}
\mathbf{K} = 
\begin{bmatrix}
f_x & 0   & c_x \\
0   & f_y & c_y \\
0   & 0   & 1
\end{bmatrix}
\end{equation}
Extrinsic parameters describe the camera's position and orientation in 3D space relative to the scene (world coordinates) and are made of the translation vector $\mathbf{t} \in \mathbb{R}^{1 \times 3}$ and the rotation matrix $\mathbf{R} \in \mathbb{R}^{3 \times 3}$. These are used to bring the world origin into camera's origin and to rotate the world coordinate system into the camera's orientation. The complete projection equation from 3D world coordinates to 2D image coordinates used for our Gaussian projection is formulated:
\begin{equation}
\mathbf{x}_\text{image} \sim \mathbf{K} [\mathbf{R} \mid \mathbf{t}] \mathbf{X}_\text{world}
\end{equation}
where $\mathbf{x}_\text{image}$ and $\mathbf{X}_\text{world}$ are the image and world point coordinates. 
\par
3D Gaussian Splatting starts from the sparse point cloud generated by SfM and initializes a set of 3D Gaussians. By optimization, they are refined to better capture the geometry and appearance of the scene, resulting in a densified set of Gaussians. These are parameterized by their position $\mathbf{\mu} \in \mathbb{R}^3$, scale, orientation (represented through a quaternion), opacity and color. 3D-GS also uses an efficient rasterization technique for generating 2D views. First, all Gaussians are projected onto the image plane, the screen is split into 16x16 tiles and the Gaussians are distributed to the tiles they are intersecting. Next, the Gaussians are sorted based on depth and a separate thread for each tile is launched resulting in fast parallel processing. Then, for each pixel in the tile, the sorted Gaussian list is traversed from front to back. The accumulated opacity, $\alpha_{\text{acc}}$, is updated with the opacity $\alpha_i$ of each subsequent Gaussian $i$ until saturation is reached. This process can be formulated as:
\begin{equation}
\begin{aligned}
\alpha_{\text{acc}} &\leftarrow \alpha_{\text{acc}} + (1 - \alpha_{\text{acc}}) \cdot \alpha_i \\
\text{if } \alpha &\geq T, \text{ stop}
\end{aligned}
\label{eq:alpha_accumulation}
\end{equation}
where $T$ is a chosen threshold, usually 1.
Finally, a backward pass is performed, from the last Gaussian that contributed to the pixel to the front, and alpha-blending is used:
\begin{equation}
C_{\text{out}} \leftarrow \alpha_i C_i + (1 - \alpha_i) C_{\text{prev}}
\label{eq:color_blending}
\end{equation}
in order to compute the final pixel's color $C_\text{out}$. Here, $C_i$ represents the color of the current Gaussian, and $C_\text{prev}$ is the color accumulated from the Gaussians located behind it.

\subsection{Problem definition}
\par
Having a 2D mask from a single view, the goal is to lift it to 3D in order to produce a consistent and accurate 3D segmentation of the Gaussian Splatting.
Our approach uses an algorithm that projects the shapes of Gaussians in 2D, keeps those that fall inside mask boundaries and cumulates their weights in a Z-buffer. This way, the Gaussians in the foreground will progressively cover the outliers from the background. Additionally, we do a statistical filtering based on normal distributions of opacity and depth in order to further remove noise from floating Gaussians or any remaining background outliers.

\subsection{2D to 3D mask up-lifting}
\par
We will first project the centers of the Gaussians on the image plane and filter out those that fall outside the mask's boundaries. Considering a 3D point $\mathbf{X}_\text{world} = (X_w, Y_w, Z_w)$ in world coordinates, its coordinates in the camera frame $\mathbf{X}_\text{cam} = (X_c, Y_c, Z_c)$ can be formulated as:
\begin{equation}
\mathbf{X}_\text{cam} = \mathbf{R}(\mathbf{X}_\text{world}-\mathbf{t}),
\end{equation}
where $\mathbf{t} \in \mathbb{R}^{3}$ and $\mathbf{R} \in \mathbb{R}^{3 \times 3}$ are the camera's translation vector and rotation matrix, respectively. To project this 3D camera point into 2D image coordinates $(x, y)$, we use the following equations:
\begin{equation}
x = f_x \cdot \frac{X_c}{Z_c} + \frac{w}{2},\,\,\,\,\,\,\,y = f_y \cdot \frac{Y_c}{Z_c} + \frac{h}{2},
\end{equation}
where $(f_x, f_y)$ are the camera's focal lengths and $(w, h)$ is the image size in pixels.

\par
Having the Gaussians inside the 2D mask, we will start the Z-buffering algorithm by sorting them in ascending order of the distance from the camera.
For each Gaussian, we will check if its center position in the buffer is lower than $\boldsymbol{\beta}$, a threshold that is calculated as the mean opacity of the Gaussians already in the buffer. 
\par
For the 2D shape projection we begin with a 3D Gaussian represented by a mean position $\mathbf{p} \in \mathbb{R}^3$, a rotation (orientation) matrix $\mathbf{R}_\text{g} \in SO(3)$ derived from a quaternion, and a scale vector $\mathbf{s} \in \mathbb{R}^3$. The scale is stored in logarithmic form for numerical stability, so we define the scale matrix as:
\begin{equation}
\boldmath
S = \text{diag}(\exp(s_x), \exp(s_y), \exp(s_z))
\end{equation}
The 3D covariance matrix of the Gaussian in world space is constructed by rotating the scaled ellipsoid:
\begin{equation}
\boldsymbol{\Sigma}_\text{world} = \mathbf{R}_\text{g} \mathbf{S}^2 \mathbf{R}_\text{g}^\top
\end{equation}
To project this Gaussian into the camera frame, we apply the camera's rotation matrix $\mathbf{R}_\text{cam}$:
\begin{equation}
\boldsymbol{\Sigma}_\text{cam} = \mathbf{R}_\text{cam} \boldsymbol{\Sigma}_\text{world} \mathbf{R}_\text{cam}^\top
\end{equation}
Next, we project the 3D covariance into 2D image space using the Jacobian of the perspective projection. For a point $\mathbf{p}_\text{cam} = (x, y, z)$ in camera coordinates, the Jacobian $\mathbf{J}$ is:
\begin{equation}
\mathbf{J} = 
\begin{bmatrix}
\frac{f_x}{z} & 0 & -\frac{f_x x}{z^2} \\
0 & \frac{f_y}{z} & -\frac{f_y y}{z^2}
\end{bmatrix}
\end{equation}
where $f_x$ and $f_y$ are the camera focal lengths in pixels. The 2D image-space covariance becomes:
\begin{equation}
\boldsymbol{\Sigma}_\text{image} = \mathbf{J} \boldsymbol{\Sigma}_\text{cam} \mathbf{J}^\top
\end{equation}
This covariance describes an elliptical Gaussian in the 2D image. To evaluate the spatial extent of this Gaussian, 
we compute the Mahalanobis distance for each pixel within a region of interest:
\begin{equation}
D^2(\mathbf{x}) = (\mathbf{x} - \bm{\mu})^\top \boldsymbol{\Sigma}_\text{image}^{-1} (\mathbf{x} - \bm{\mu})
\end{equation}
where $\bm{\mu} = (x_\text{proj}, y_\text{proj})$ is the projected mean. The final Gaussian weight at each pixel is computed as:
\begin{equation}
w(\mathbf{x}) = \alpha \cdot \exp\left(-\frac{1}{2} D^2(\mathbf{x})\right)
\end{equation}
where $\alpha$ is the opacity. This weight value represents the contribution of the Gaussian to that pixel.

\par
As previously mentioned, the weights are summed into a buffer resulting in a subset of 3D Gaussians. Artifacts caused by poor reconstruction quality may be present in the scene. Thus, we remove Gaussians that do not fall within two standard deviations (2$\sigma$) of the mean depth and opacity values of the current subset.

\begin{figure}[htbp]
\centering
\includegraphics[width=0.9\linewidth]{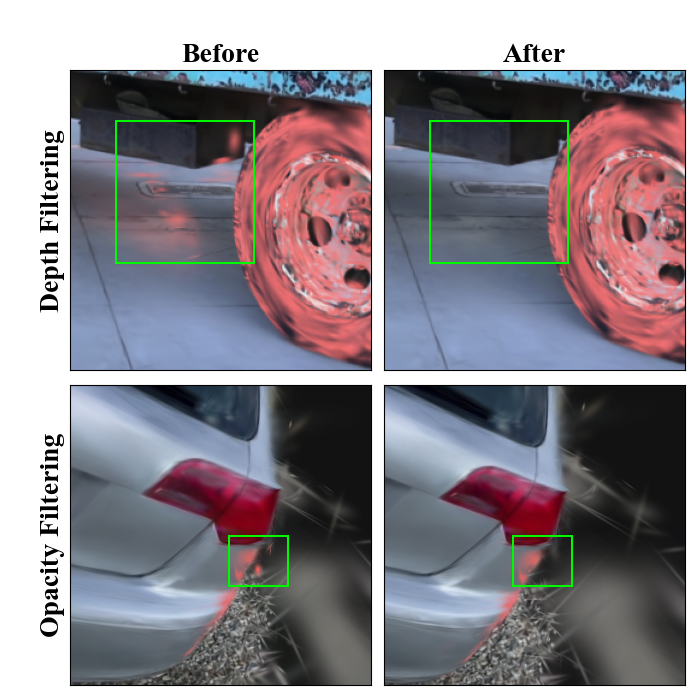}
\caption{The effect of statistical filtering on segmentation quality. Our method removes background outliers based on depth and eliminates noisy floating Gaussians based on opacity.}
\label{fig:filter}
\end{figure}

Fig.~\ref{fig:filter} shows us a comparison of segmentations with and without the statistical filtering for opacity and depth. In case of depth, we can see that there are visible background Gaussians that were improperly segmented, mostly because of failures in the Z-buffering algorithm. These outliers are successfully filtered out by using the normal distribution of depth and only keeping Gaussians within 2 standard deviations from the mean. The same thing is done with the opacity value. Floating Gaussians between the camera and the object caused by noise are wrongly segmented but filtered out by applying the same approach.
\par During depth-buffering, some Gaussians may be wrongly filtered out due to overlapping in the grid, causing an incomplete or sparse mask. Thus, this algorithm may not be enough for obtaining our final segmentation, but very effective for identifying the boundaries of the 3D mask by suppressing background artifacts. Consequently, we traverse again the list of Gaussian and fill the mask with those that are in between 2 standard deviations away from the mask's mean depth.

\section{Experiments and results}\label{sec:experiments}

\begin{figure*}[htbp]
    \centering
    
    % First row - single image
    \includegraphics[width=0.30\linewidth]{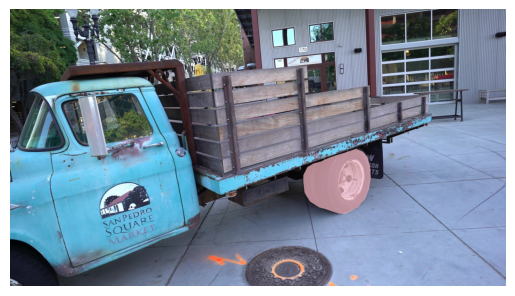}
    \includegraphics[width=0.30\linewidth]{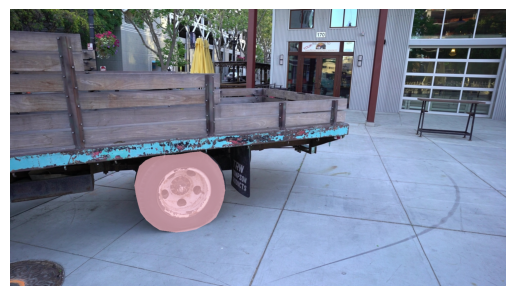}
    \includegraphics[width=0.30\linewidth]{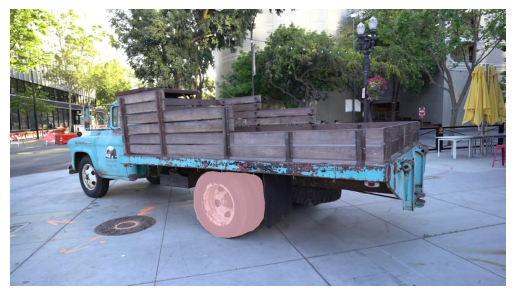}
    % \caption*{Image with Mask}

    \begin{minipage}{\linewidth}
        \centering
        \begin{minipage}{0.3\linewidth}
            \centering
            \small (a) Left-view mask
        \end{minipage}
        \begin{minipage}{0.3\linewidth}
            \centering
            \small (b) Center-view mask
        \end{minipage}
        \begin{minipage}{0.3\linewidth}
            \centering
            \small (c) Right-view mask
        \end{minipage}
    \end{minipage}
    
    \vspace{1em} % Add some vertical space between rows
    
    % Second row - three images side by side
    \includegraphics[width=0.16\linewidth]{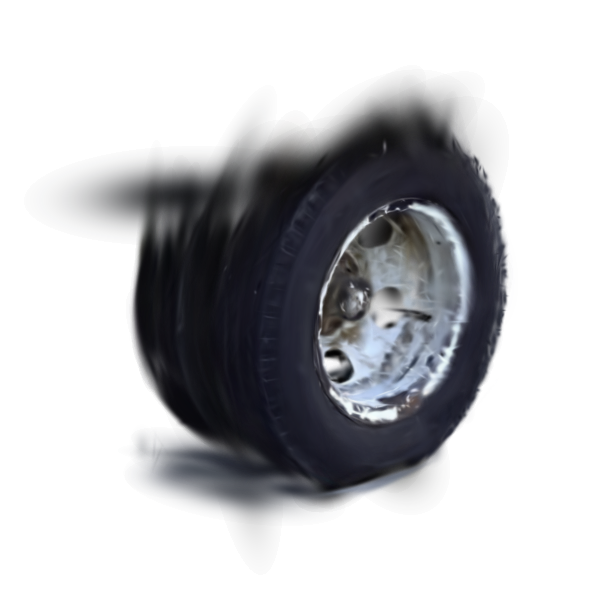}
    \includegraphics[width=0.16\linewidth]{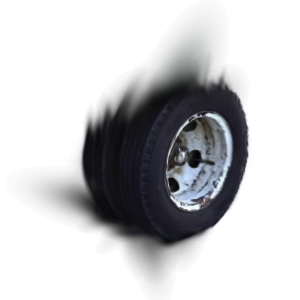}
    \includegraphics[width=0.16\linewidth]{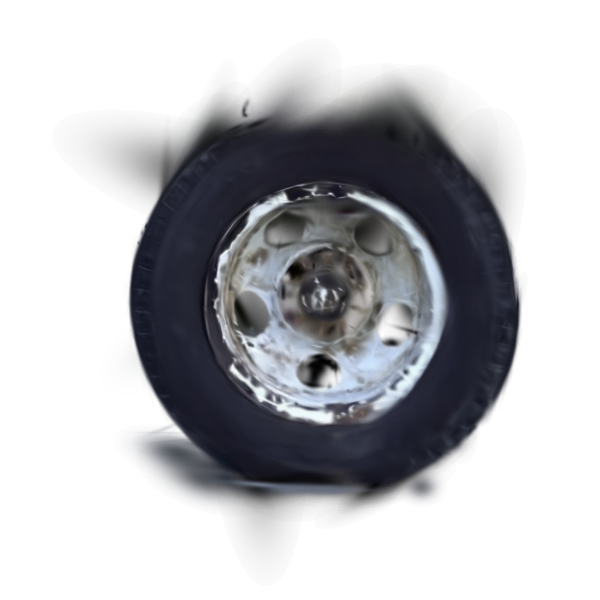}
    \includegraphics[width=0.16\linewidth]{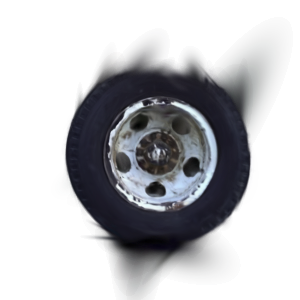}
    \includegraphics[width=0.16\linewidth]{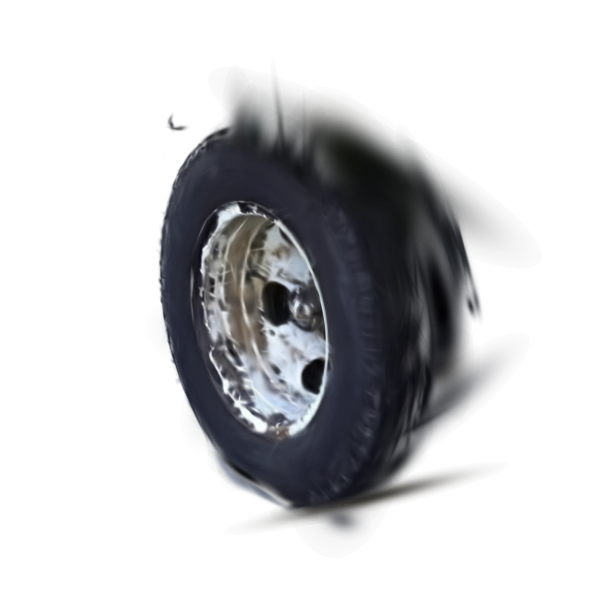}
    \includegraphics[width=0.16\linewidth]{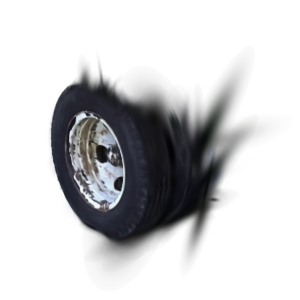}
    
    % Simple text captions without "Fig." prefix
    \begin{minipage}{\linewidth}
        \centering
        \begin{minipage}{0.12\linewidth}
            \centering
            \small Ours(left)
        \end{minipage}
        \begin{minipage}{0.19\linewidth}
            \centering
            \small SAGD(left)
        \end{minipage}
        \begin{minipage}{0.15\linewidth}
            \centering
            \small Ours(center)
        \end{minipage}
        \begin{minipage}{0.21\linewidth}
            \centering
            \small SAGD(center)
        \end{minipage}
        \begin{minipage}{0.15\linewidth}
            \centering
            \small Ours(right)
        \end{minipage}
        \begin{minipage}{0.15\linewidth}
            \centering
            \small SAGD(right)
        \end{minipage}
    \end{minipage}
    
    \caption{Qualitative comparison with the multi-view method SAGD~\cite{HuXu2025}. SAGD's approach requires consistent masks from multiple views (a, b, c) to produce its segmentation. In contrast, our method achieves a visually comparable result using only a single mask from the center view (b), demonstrating its effectiveness for scenarios where obtaining reliable multi-view annotations is impractical or impossible.}
    \label{fig:wheel_comparison}
\end{figure*}

\subsection{2D Car damage segmentation}\label{sec:41}
For the instance segmentation of car damage, we conducted experiments on both existing car damage datasets, namely CarDD~\cite{WangXinkuang2023} and VehiDE~\cite{HuynhNhan2023}. These datasets contain segmentation labels for various damage classes. Table~\ref{tab:damage-datasets} presents a comparison including the number of images, instances and damage categories of these datasets.

\renewcommand{\arraystretch}{1.4} % Increase row height for better readability
\begin{table}[h]
\centering
\caption{Overview of publicly available car damage detection datasets.}
{\footnotesize
\begin{tabular}{l c c p{3.5cm} } % last column uses fixed-width paragraph column
\hline
\textbf{Dataset} & \textbf{Images} & \textbf{Instances} & \textbf{Damage Categories} \\
\hline
CarDD\cite{WangXinkuang2023} & 4000 & 9000 & scratch, dent, glass shatter, crack, tire flat, and lamp broken \\
\hline
VehiDE\cite{HuynhNhan2023} & 13945 & 32000 & dents, broken glass, scratch, lost parts, punctured, torn, broken lights, and non-damage \\
\hline
\end{tabular}
}
\label{tab:damage-datasets}
\end{table}

\par In both dataset papers, authors introduce experiments with different Recurrent Convolutional Neural Network architectures and conclude that the best result is obtained by DCN (Deformable convolutional networks)~\cite{DaiJifeng2017}. Consequently, we decided to run experiments using a different type of architecture, namely YOLO~\cite{RedmonJoseph2016}. We trained two versions of the YOLOv11 network from Ultralytics~\cite{JocherGlenn2023}.The datasets train/test split was made using the original structure provided by the
authors. CarDD has 2816 samples in the training set and 810 in the validation set. VehiDE provides 11621 training samples and 2324 validation samples, with the rest of the samples being in the test sets.

\begin{table}[h]
\centering
\caption{Car damage instance segmentation training results in terms of box (b) and mask (m) Average Precision.}
\label{table:comparison}
{\footnotesize
\begin{tabular}{l c c c c c c c c}
\hline
\textbf{Dataset} & \textbf{Network} & $mAP50$ (b/m) & $mAP50-95$ (b/m) \\
\hline
\multirow{2}{*}{CarDD\cite{WangXinkuang2023}} & YOLO11-l & 76.7 / 75.8 & 61.8 / 58.5 \\
& YOLO11-x & 76.2 / 75.5 & 62.0 / 58.6 \\
\hline
\multirow{2}{*}{VehiDE\cite{HuynhNhan2023}} & YOLO11-l & 53.9 / 51.2 & 36.1 / 29.4 \\
& YOLO11-x & 55.4 / 52.3 & 37.5 / 30.2 \\
\hline
\end{tabular}
}
\end{table}
\par Table~\ref{table:comparison} shows a summary of mAP metrics for both datasets and YOLO models. In summary, the YOLO11-x version performs slightly better than YOLO11-l, but having more than double parameters, making us choose the lighter version. The inference time for YOLO11-l on our system (Apple M3 Pro) is under 100ms for one image of size 640x640.

\subsection{3D Gaussian Splatting Segmentation}
\label{subsection1}

\par
In order to demonstrate the robustness of our method, we conducted experiments on self-recorded data of damaged cars but also on samples from public 3D-GS datasets.

\par In order to check the consistency of our projected mask, we run the algorithm and back-project the segmentation result onto the original image that was used as input for the segmentation. The resulting 2D mask is a set of 2D Gaussian shapes that may have an irregular boundary or may not be fully connected. In order to compare this mask with the ground truth one, we use convex and concave hulls, depending on the resulting shape (we use this step only for the comparisons on self-recorded data). Now, the generated polygon is compared with ground truth in terms of Intersection over Union, F1 score and Accuracy. Some qualitative results of this experiment are shown in Fig.~\ref{fig:iou}. We demonstrate the consistency of our segmentation results by visualizing the generated masks from different angles, on the rendering and projected on vehicle images.
\par

\begin{figure*}[ht]
\centering
\setlength{\tabcolsep}{2pt} % spacing between images
\renewcommand{\arraystretch}{1.0}
\begin{tabular}{c cccc} % 3 columns
    % Row 1
    % Singe view mask & fork & lego & pinecone \\  % column labels
    \rotatebox{90}{\raisebox{.5\height}{fortress}}
    \includegraphics[width=0.2\textwidth]{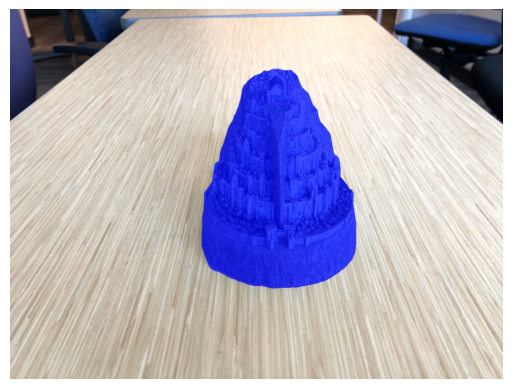} &
    \includegraphics[width=0.2\textwidth]{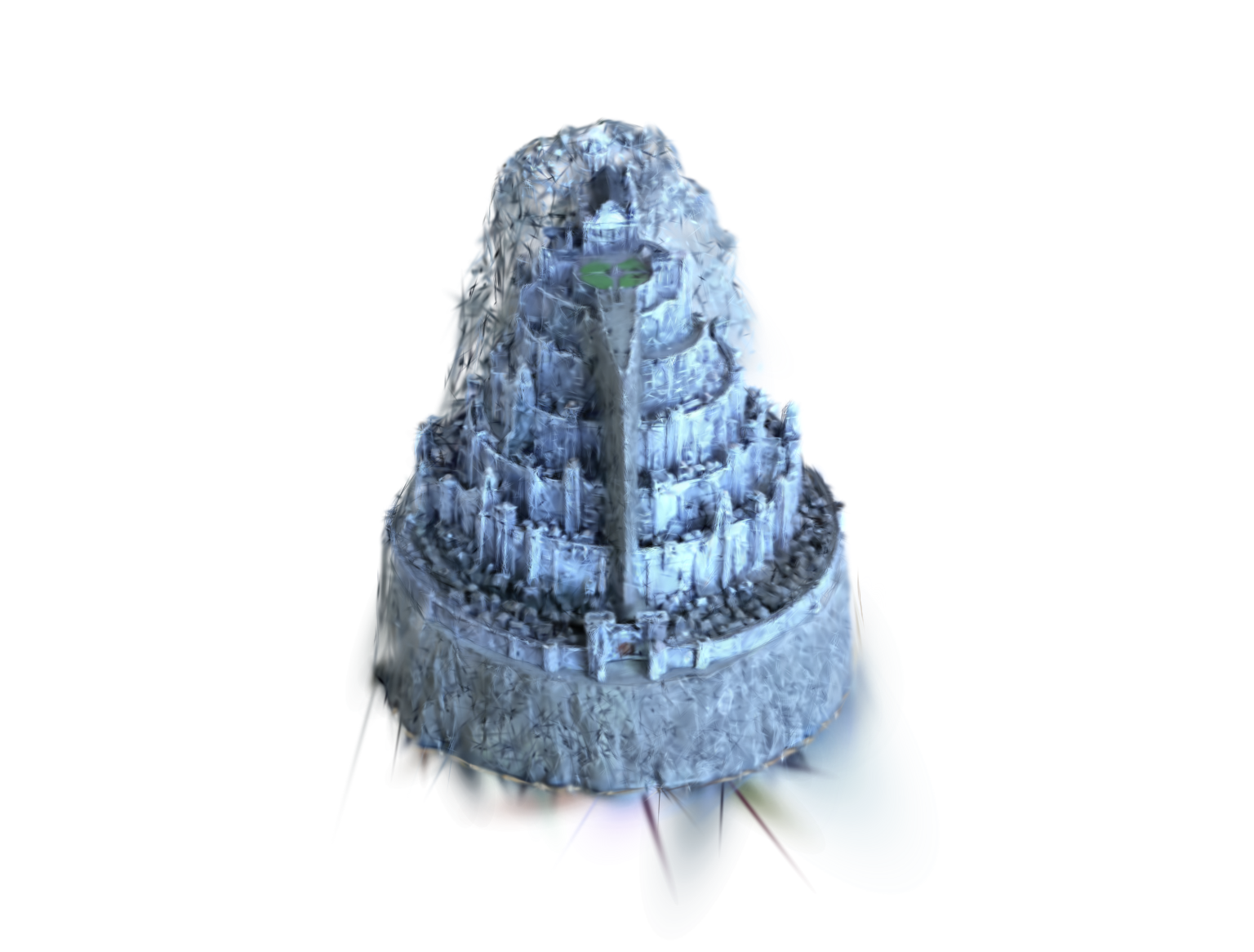} &
    \includegraphics[width=0.2\textwidth]{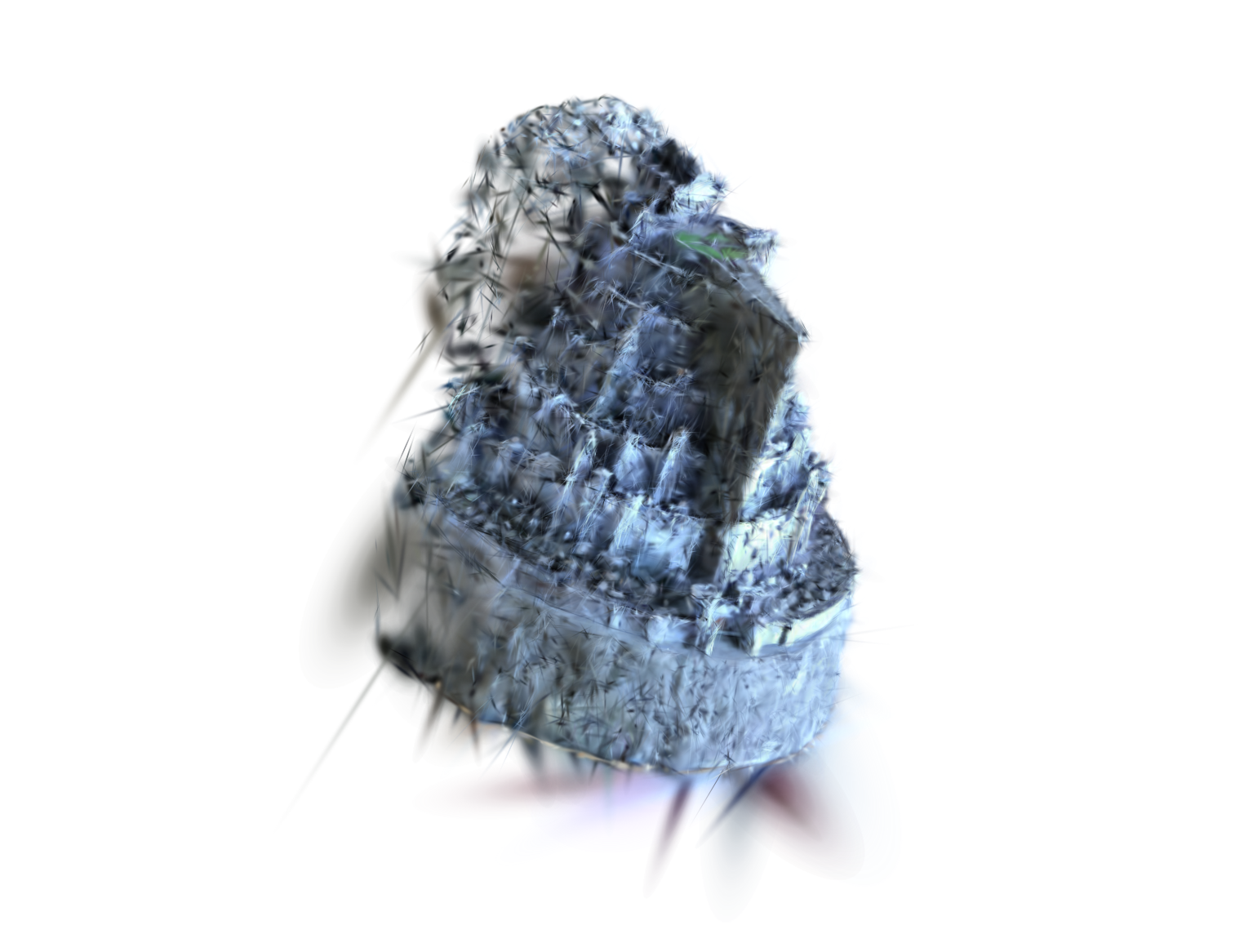} &
    \includegraphics[width=0.2\textwidth]{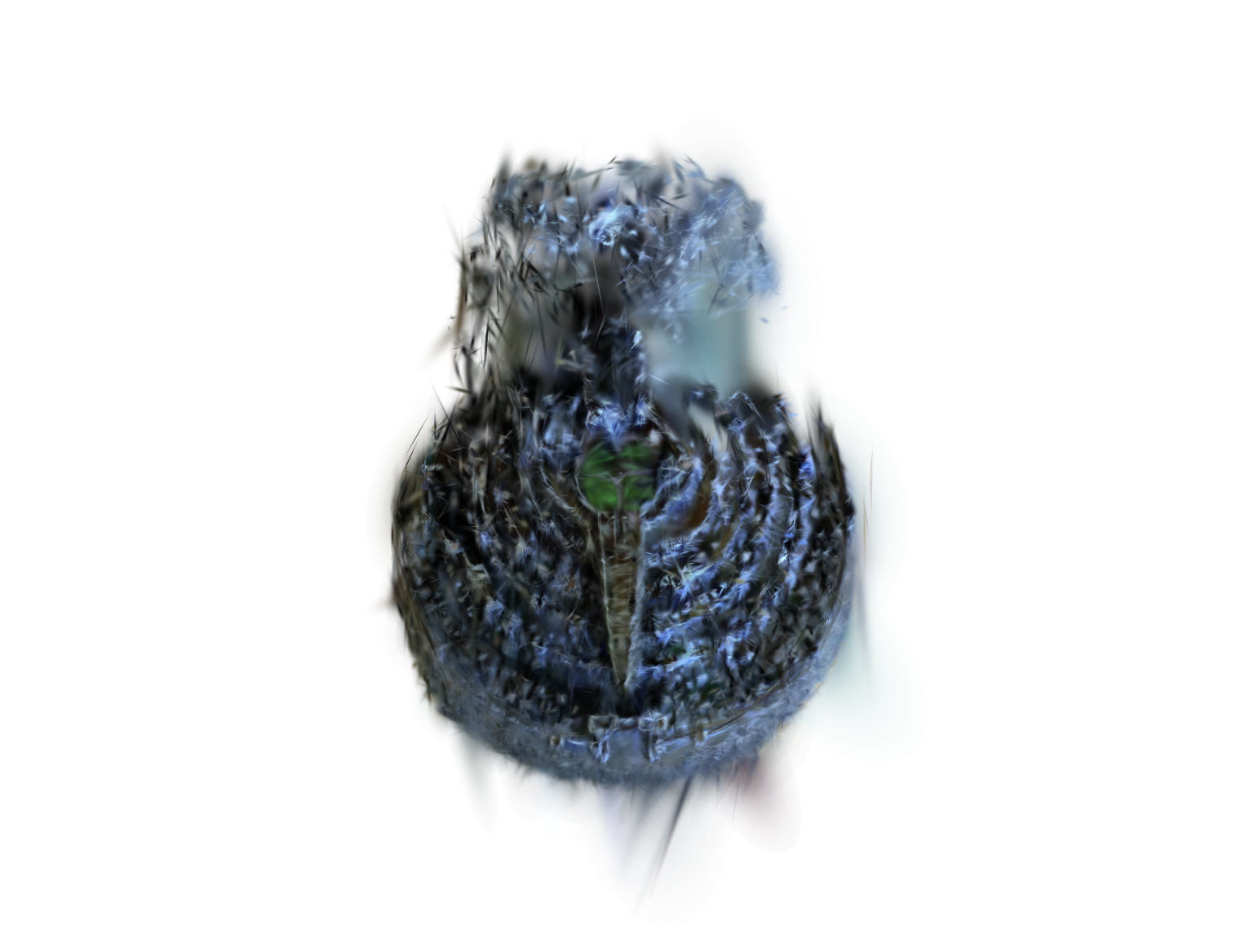} \\
    \rotatebox{90}{\raisebox{.5\height}{lego}}
    % Row 2
    \includegraphics[width=0.2\textwidth]{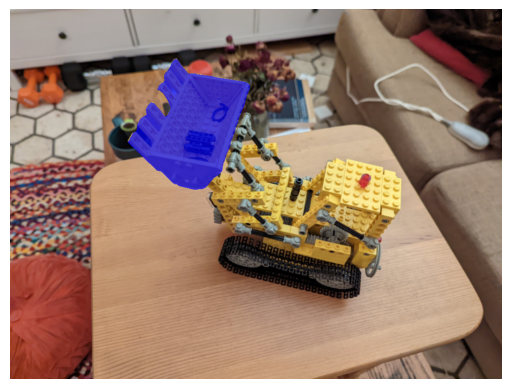} &
    \includegraphics[width=0.2\textwidth]{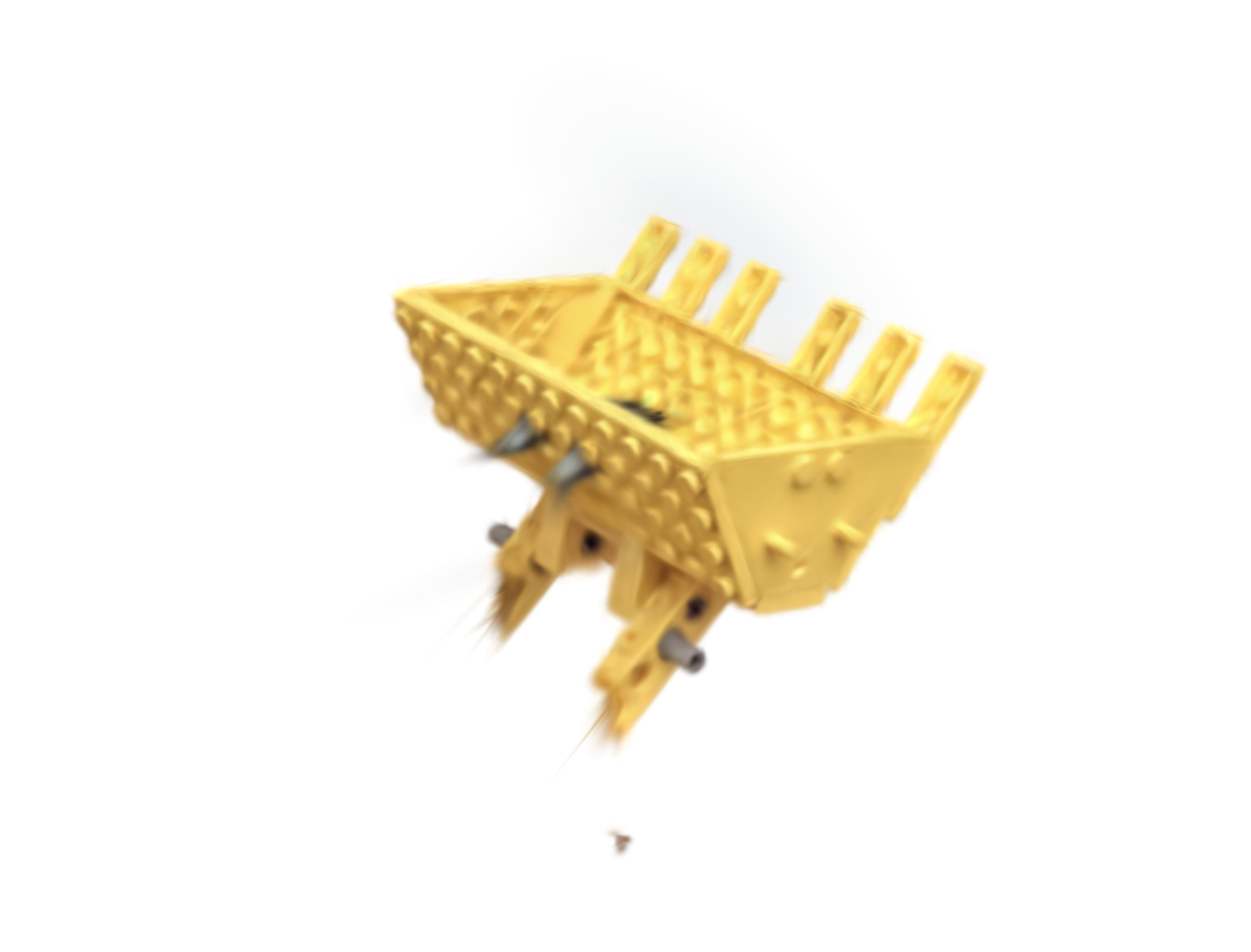} &
    \includegraphics[width=0.2\textwidth]{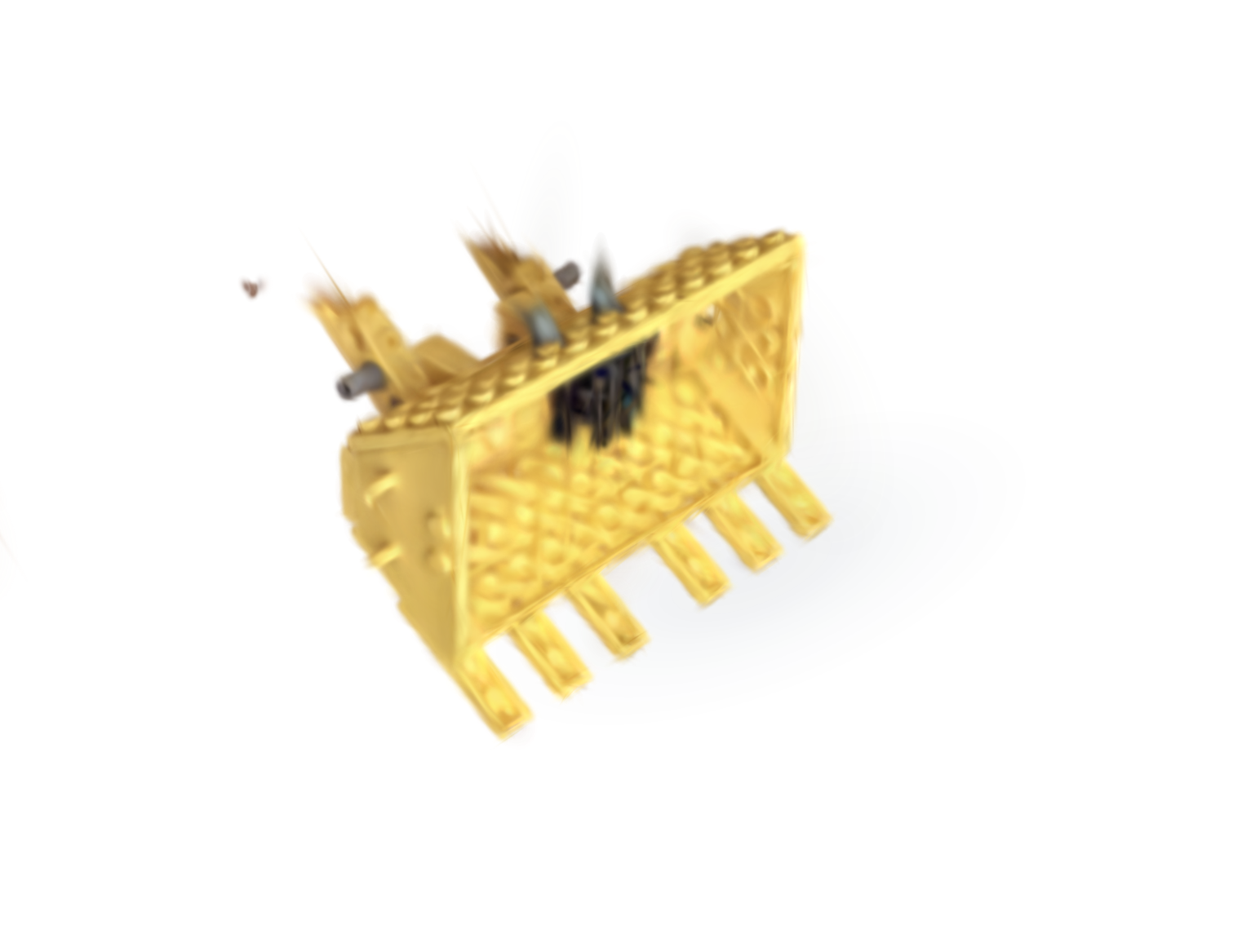} &
    \includegraphics[width=0.2\textwidth]{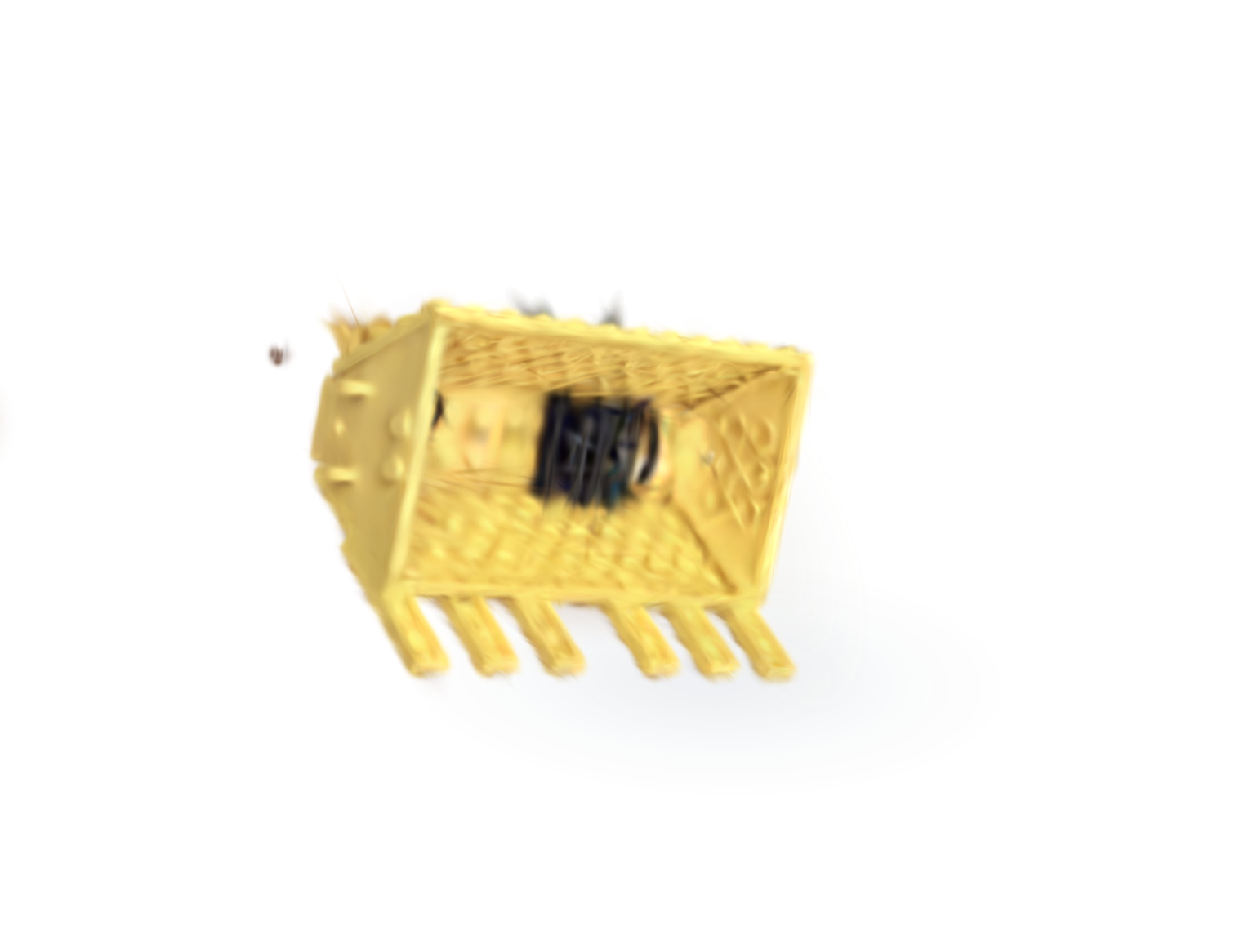} \\
    \rotatebox{90}{\raisebox{.5\height}{orchids}}
    % Row 3
    \includegraphics[width=0.2\textwidth]{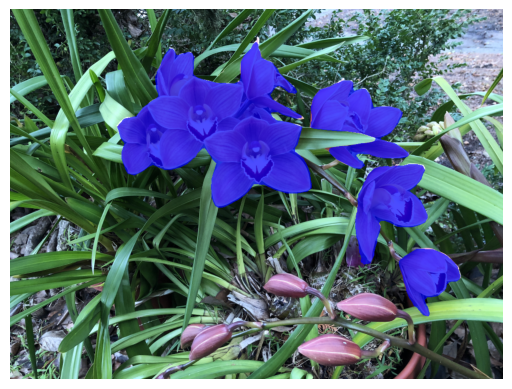} &
    \includegraphics[width=0.2\textwidth]{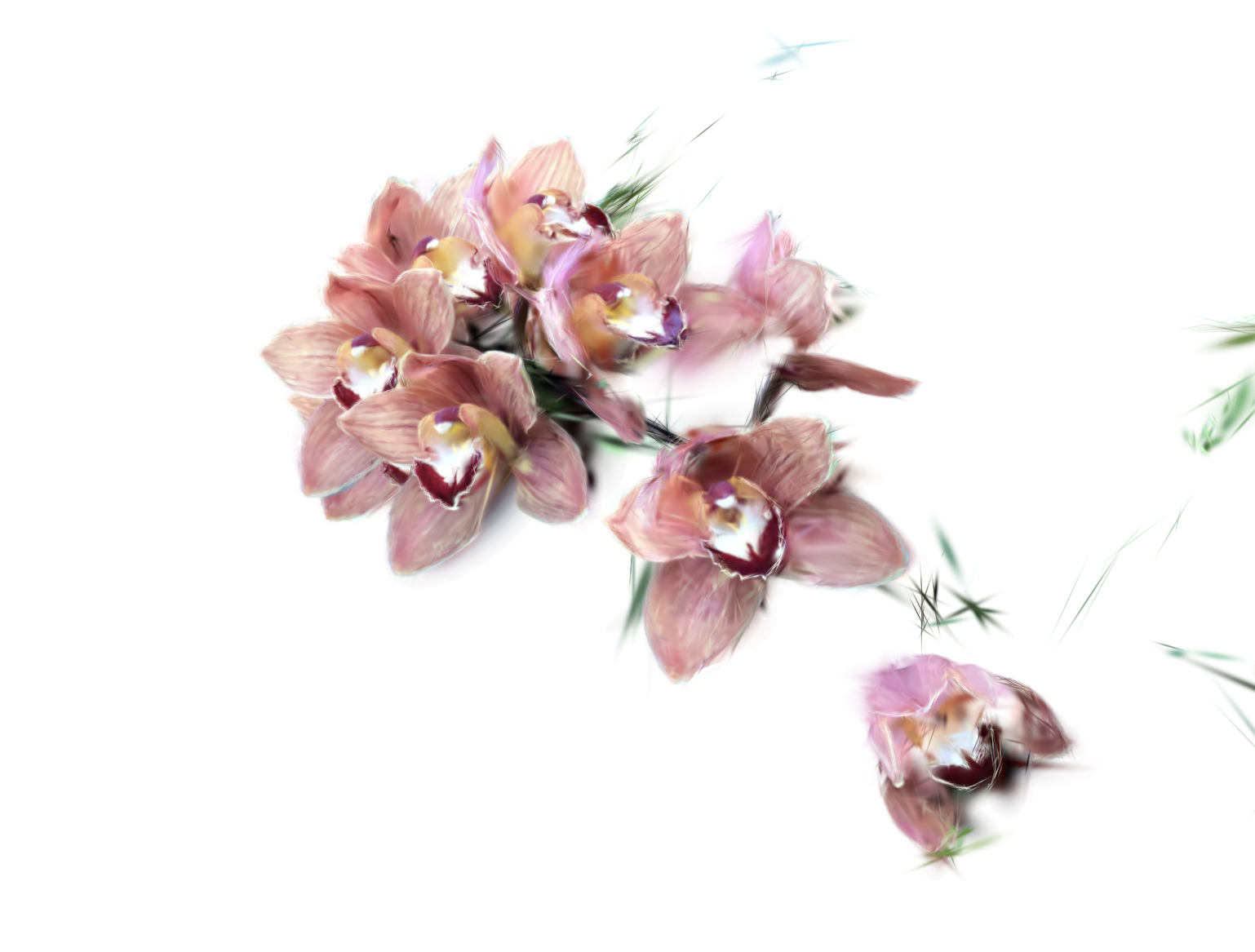} &
    \includegraphics[width=0.2\textwidth]{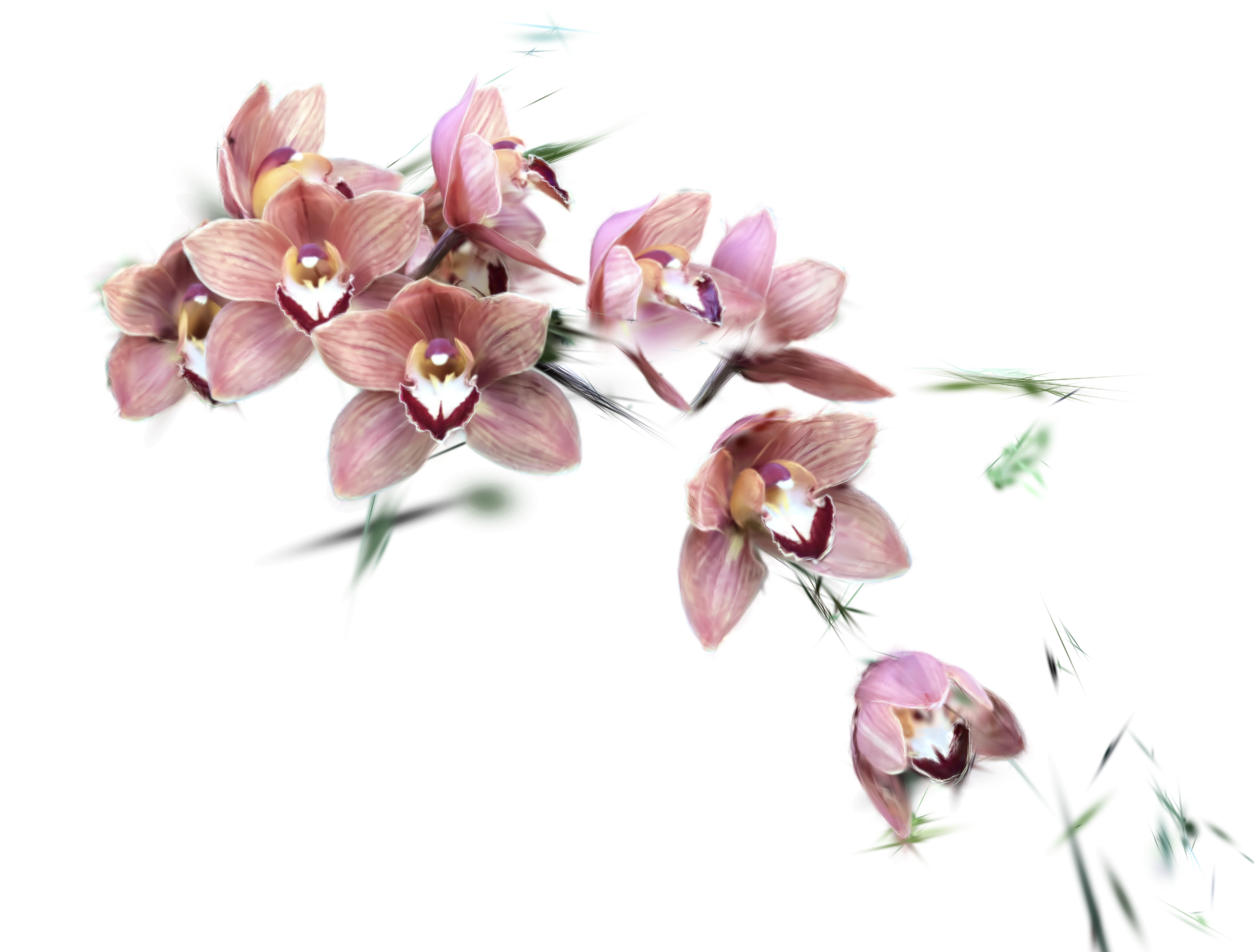} &
    \includegraphics[width=0.2\textwidth]{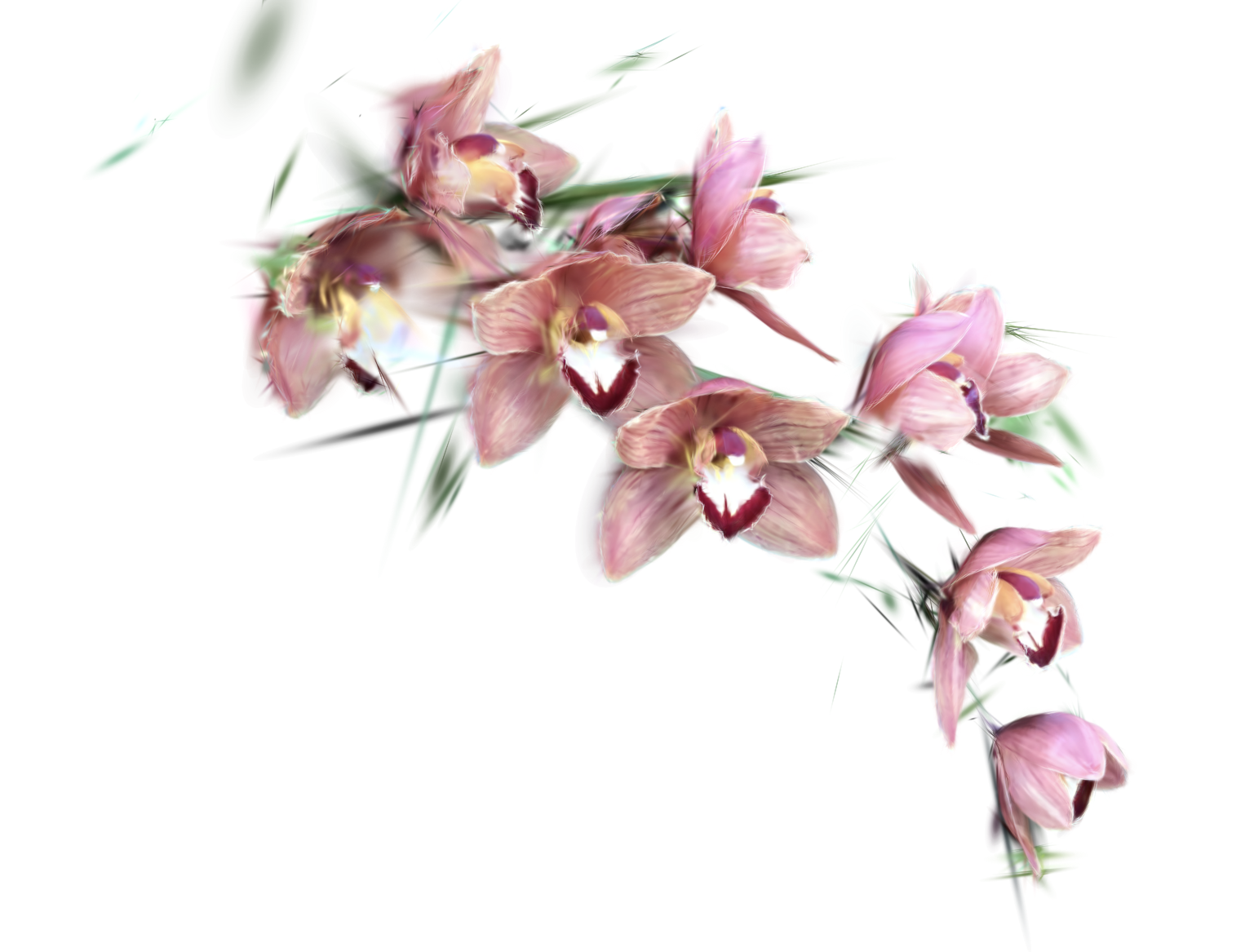} \\
    \rotatebox{90}{\raisebox{.5\height}{pinecone}}
    % Row 4
    \includegraphics[width=0.2\textwidth]{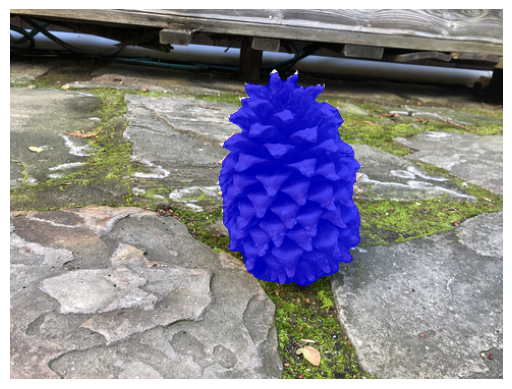} &
    \includegraphics[width=0.2\textwidth]{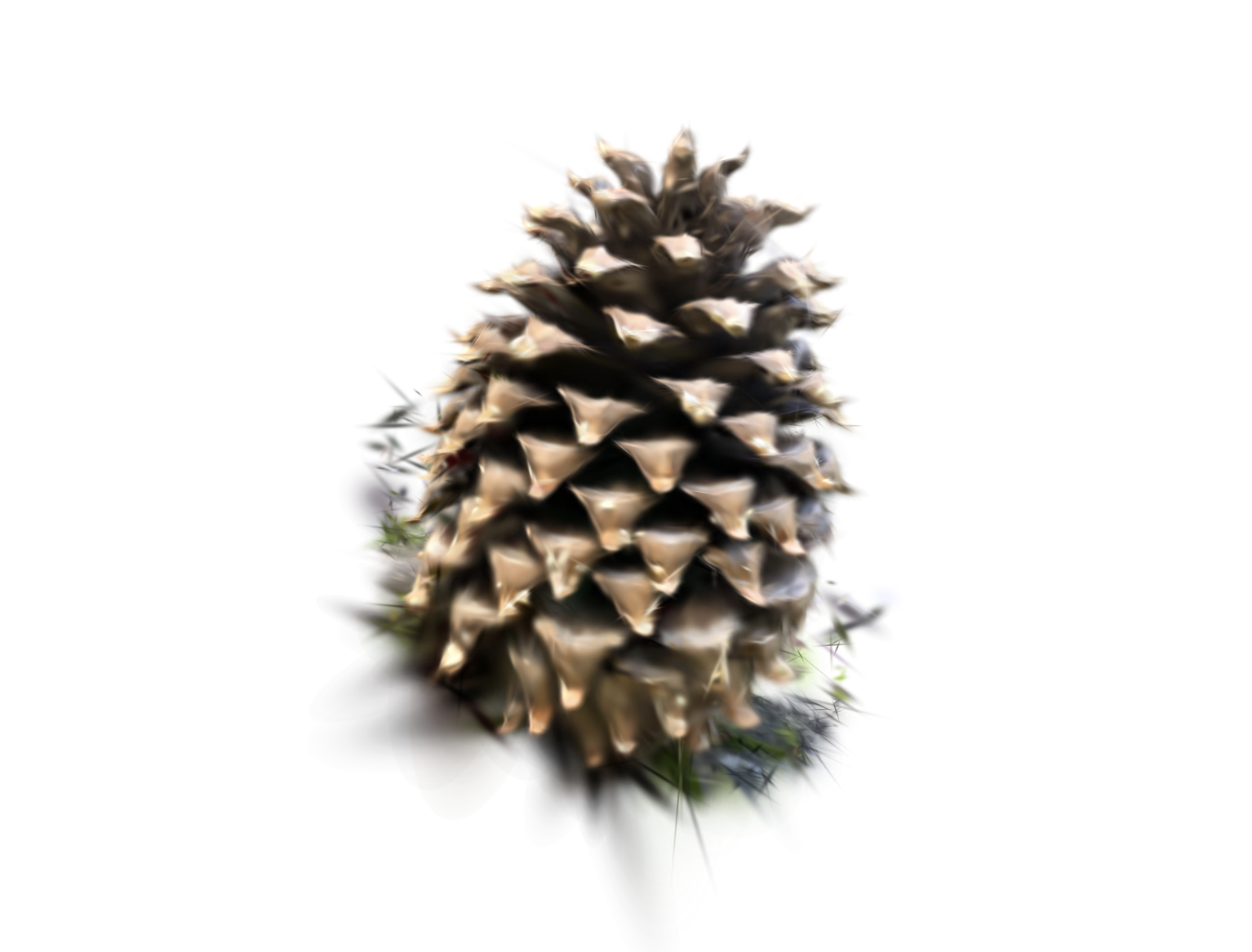} &
    \includegraphics[width=0.2\textwidth]{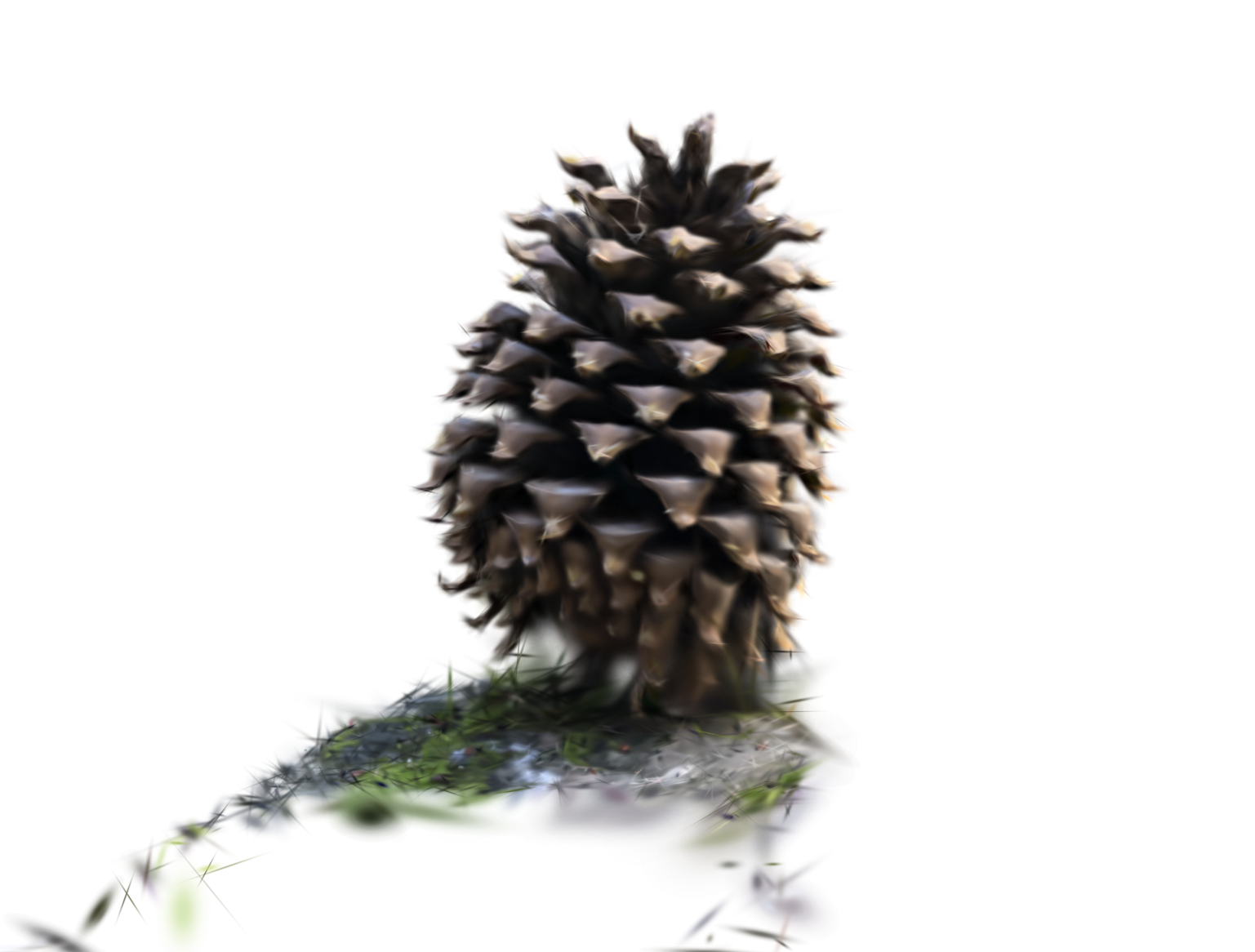} &
    \includegraphics[width=0.2\textwidth]{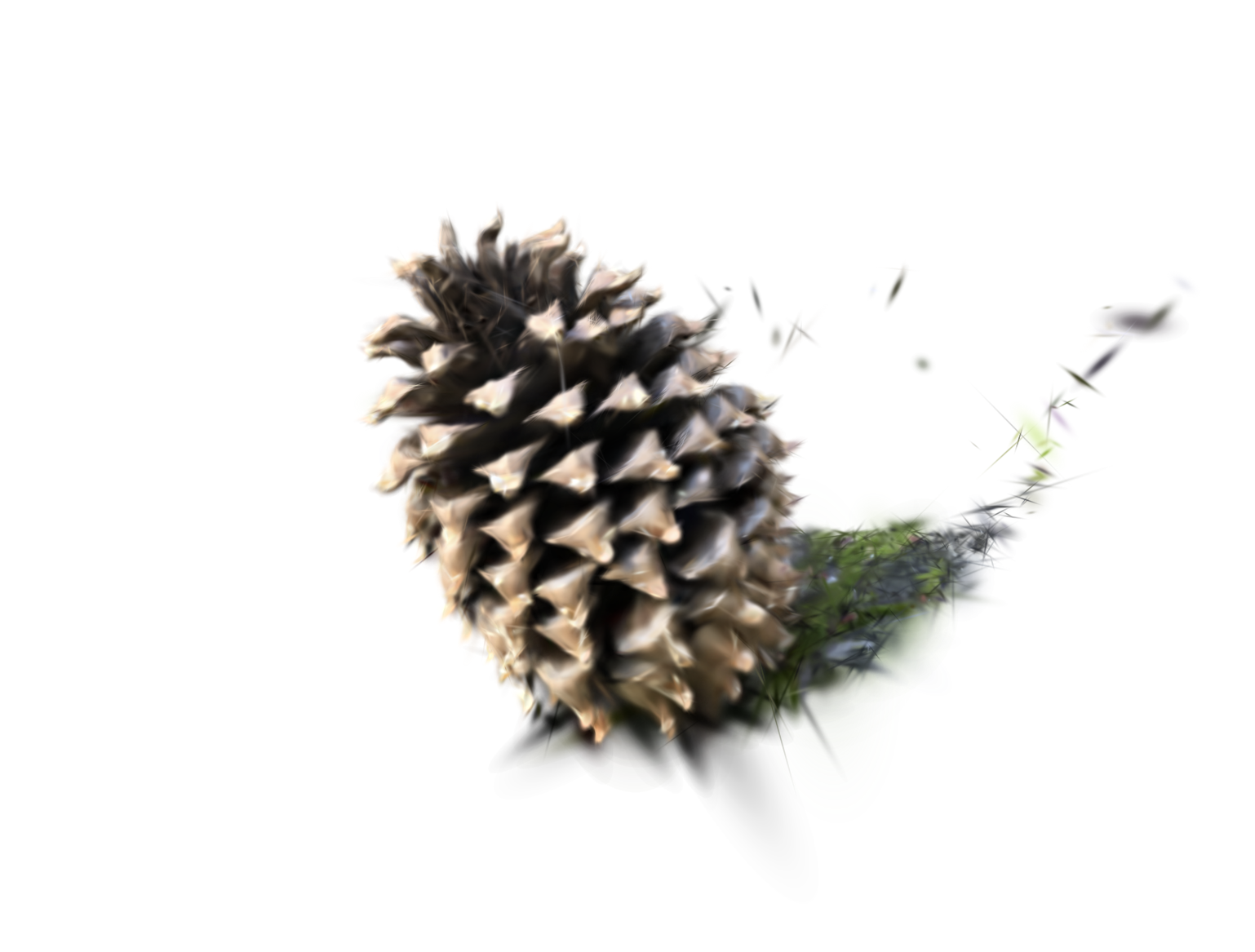} \\
\end{tabular}

\caption{Qualitative comparison of our method on scenes from SPIn-NeRF~\cite{MirazaeiAshkan2023}.}
\label{fig:spin_vcomparison}
\end{figure*}

The mentioned metrics used for the comparison are presented in Table~\ref{table:quant_comparison}, together with the processing time required for the segmentation. Besides the comparison with the view used as input, we manually labeled segmentation masks for two more views that capture the damage, (g) and (h) from Fig.~\ref{fig:iou} (the masks displayed in the figures are the projection of our result, not these ground truth masks). We calculate the mean metrics over these three views in order to show that our results are consistent across multiple unseen views.
\par
In order to compare our method with other studies, we use the SPIn-NeRF~\cite{MirazaeiAshkan2023} which include 2D segmentation masks of some objects from multiple public 3D reconstruction datasets. For the evaluation, we use only one ground truth mask that we up-lift using our method. The resulting 3D segmentation is then projected in all the views from the scene and compared with the ground truth mask in order to compute IoU and Accuracy. These results are compared with other 3D segmentation methods in Table~\ref{tab:segmentation_results}. The results reported for the other methods are taken directly from the original publications, namely SA3D~\cite{CenJiazhong2025_2} and SAGD~\cite{HuXu2025}. Also, for generating the 3DGS scenes, we used OpenSplat~\cite{ToffaninPiero2024} and optimized for 7k steps. The underline results mark the scenes where our single-view approach outperformed multi-view methods.
\par We could observe that although only one view was used for the segmentation, the results are comparable with multi-view approaches, some of them requiring scene-specific training. More than this, we also outperformed the baseline multi-view segmentation method (Single View~\cite{CenJiazhong2025_2}) on some scenes, obtaining a similar mean for the IoU and Accuracy on the SPIn-NeRF scenes. Despite its name, this baseline method uses a multi-view approach. It starts from an initial ground-truth mask that mapped to the 3D space using depth information in order to obtain an initial 3D segmentation. This result is then back-projected into other views, resulting in 2D prompt points that are passed to SAM in order to further generate 2D masks into the corresponding views.

\begin{table}[h]
\centering
\caption{Quantitative results of our segmentation method. All metrics are calculated compared to the input ground truth masks used for segmentation - left / mean across three different views (the input one and two unseen views) - right.}
\label{table:quant_comparison}
{\scriptsize
\begin{tabular}{l c c c c}
\hline
\textbf{Object} & \textbf{IoU(\%)} & \textbf{F1(\%)} & \textbf{Acc(\%)} & \textbf{Time(s)} \\
\hline
\makecell{Scratch \\ (Fig.~\ref{fig:iou} - top)} & 65.69 / 52.44 & 79.29 / 67.82 & 99.30 / 99.28 & 0.04 \\
\hline
\makecell{Flat tire \\ (Fig.~\ref{fig:iou} - center)} & 88.15 / 87.16 & 93.70 / 93.13 & 99.01 / 99.01 & 0.12 \\
\hline
\makecell{Broken lamp \\ (Fig.~\ref{fig:iou} - bottom)} & 82.41 / 67.07 & 90.36 / 78.75 & 97.93 / 97.70 & 0.31 \\
\hline
\end{tabular}
}
\end{table}

Another important comparison that we conduct is on the commonly used "Truck" scene from the Tanks and Temples~\cite{KnapitschArno2017} benchmark dataset. This scene is presented in the visual results from most of the 3D-GS segmentation studies. In Fig.~\ref{fig:wheel_comparison} we compare our method with SAGD~\cite{HuXu2025}, which introduces a very intuitive learning-free segmentation approach based on a multi-view label voting mechanism. Our method obtained similar results by using only a single view for the segmentation and without any needed preprocessing.

% \begin{table}[h]
% \centering
% \caption{Running time for our method compared to SAGD.}
% \label{table:sagd_comparison}
% {\scriptsize
% \begin{tabular}{l c c c c}
% \hline
% \textbf{Method} & \textbf{Hardware} & \textbf{Input} & \makecell{\textbf{Pre-processing} \\ \textbf{overhead}} & \makecell{\textbf{Segmentation} \\ \textbf{Time(s)}} \\
% \hline
% \makecell{CrashSplat\\(ours)} & CPU & \makecell{Single \\ 2D mask} & \makecell{YOLO inference \\ (fast)} & 2.6 \\
% \hline
% SAGD~\cite{HuXu2025} & GPU & \makecell{Multiple \\ 2D masks} & \makecell{SAM inference \\ on N views (slow)} & 0.3 \\
% \hline
% \end{tabular}
% }
% \end{table}

\begin{table*}[ht]
\centering
\caption{Quantitative evaluation on SPIn-NeRF~\cite{MirazaeiAshkan2023} scenes, comparing our method with existing multi-view approaches. Underlined entries highlight scenarios where our single-view method outperforms the multi-view baselines (Single View~\cite{CenJiazhong2025_2} is a multi-view method despite its name (see Section~\ref{subsection1}).}
\setlength{\tabcolsep}{6pt} % adjust column spacing
\renewcommand{\arraystretch}{1.2} % adjust row spacing
\begin{tabular}{>{\centering\arraybackslash}m{2.2cm}|cc|cc|cc|cc|cc}
\toprule
Scenes & \multicolumn{2}{c|}{Single View~\cite{CenJiazhong2025_2}} & \multicolumn{2}{c|}{MVSeg~\cite{MirazaeiAshkan2023}} & \multicolumn{2}{c|}{SA3D~\cite{CenJiazhong2025_2}} & \multicolumn{2}{c|}{SAGD~\cite{HuXu2025}} & \multicolumn{2}{c}{Ours} \\
\cmidrule(lr){2-11}
% \cmidrule(lr){4-5}\cmidrule(lr){6-7}\cmidrule(lr){8-9}\cmidrule(lr){10-11}
 & IoU & Acc & IoU & Acc & IoU & Acc & IoU & Acc & IoU & Acc \\
\midrule
Orchids   & 79.4 & 96.0 & 92.7 & 98.8 & 83.6 & 96.9 & 85.4 & 97.5 & 72.1 & 93.9 \\
Ferns     & 95.2 & 99.3 & 94.3 & 99.2 & 97.1 & 99.6 & 92.0 & 98.9 & 76.0 & 96.0 \\
Horns     & \underline{85.3} & \underline{97.1} & 92.8 & 98.7 & 94.5 & 99.0 & 91.1 & 98.4 & \underline{89.1} & \underline{98.0} \\
Fortress  & 94.1 & 99.1 & 97.7 & 99.7 & 98.3 & 99.8 & 95.6 & 99.5 & 83.2 & 97.4 \\
Pinecone  & \underline{57.0} & \underline{92.5} & 93.4 & 99.2 & 92.9 & 99.1 & 92.6 & 99.0 & \underline{81.3} & \underline{97.3} \\
Lego      & \underline{76.0} & \underline{99.1} & \underline{74.9} & 99.2 & 92.2 & 99.8 & 90.2 & 99.7 & \underline{78.0} & \underline{99.2} \\
\midrule
Mean      & 81.1 & 97.1 & 90.9 & 99.1 & 93.1 & 99.0 & 91.1 & 98.8 & 79.9 & 96.9 \\
\bottomrule
\end{tabular}
\label{tab:segmentation_results}
\end{table*}

We report end-to-end, per-instance overhead, excluding reconstruction (which all methods require), as a practical measure for comparison. Our single-view, training-free pipeline uses one 2D mask and a lightweight up-lifting step, which results in sub-second CPU latency per instance in our experiments. We don’t include kernel timing comparisons across papers, as fair benchmarking would require matching hardware setups and numbers of views.
\par For additional context, in the “Truck” wheel scene, our CPU-side steps (sorting, projection, segmentation) took under 3 seconds. In comparison, we ran SAGD~\cite{HuXu2025}, a multi-view segmentation method, with its projection and voting stages taking under a second on an A100/40GB GPU, but its total runtime depends heavily on the number of views, since it requires generating a mask per view (each usually taking around 0.4 seconds, as specified in \cite{HuXu2025}). There also exist segmentation methods capable of achieving millisecond-level performance. However, they typically require scene-specific optimization, which increases preprocessing time and reduces general applicability. We therefore emphasize the practical advantage: a computationally efficient single-view method that avoids the need for multi-view setups or reliance on large foundation models like SAM, which, in addition to high processing time, can struggle with fine-grained or niche-class segmentation.

% For a fair evaluation, we measured the running time of SAGD using one A100 GPU, the same that is used in the experiment included in the original paper. The time was measured only for the projection and the multi-view label voting function (for 3 views) as a mean over 30 iterations where the first 5 were not considered. For our method, we used the exact same projection code, and measured the time for the sorting, projection and segmentation algorithm. In comparison, all our functions' code is written using Numpy and was executed on an Apple M3 Pro system. Crucially, our single-view approach bypasses the significant pre-processing overhead required by multi-view methods. SAGD relies on generating multiple masks, using a foundation model like SAM. This process can be time-consuming (SAM inference takes around 0.4 seconds per frame as SAGD states in their paper) and, as we argue, is often not feasible for challenging, sparse object classes like vehicle damage. When considering the total time from input images to final 3D segmentation, our method is highly competitive. 
\par In Fig.~\ref{fig:spin_vcomparison}, we present qualitative results of our method on several scenes from the SPIn-NeRF dataset. Although the segmentation is performed using only a single view as input, the objects in the scenes are successfully identified and segmented. Moreover, the resulting masks maintain consistency and visual quality when rendered from multiple viewpoints, demonstrating the robustness of our approach.

\section{Discussions and limitations}

\par Our automatic vehicle damage detection solution can be implemented in multiple real-world applications. Car insurance companies could benefit from it by introducing this approach in software applications for visualizing and analyzing the vehicle 3D model in order to approximate the severity and cost of the damage. Car mechanic shops could generate interactive reports and send tailored offers based on the vehicle reconstruction generated from a video taken by customers. Car-selling websites can include this as a feature in order for the potential buyers to have the most accurate visualization of the vehicle condition.

\par
Limitations of our approach can occur in multiple modules that we rely on. First, the instance segmentation network fails frequently, either by not fully covering the damage with the mask or by giving false positive damage predictions caused by shadows or reflections. The only publicly available datasets for damage detection, that we also used in this work, have only a few thousand samples, which has been demonstrated to be insufficient for robust object detection. This limitation can be solved in the future by either using a synthetic dataset approach like~\cite{ParslovJens2024} or by enhancing the detection architecture.
\par
Regarding the 3D-GS module, the quality of our final segmentation is inherently linked to the underlying reconstruction, which depends on the quantity and quality of input images. This dependency is most apparent on challenging surfaces, such as glass, where reflections and transparency can lead to noisy or low-opacity Gaussians, causing segmentation errors. While refinement methods like those proposed by~\cite{LiCongcong2025} can mitigate these artifacts, these reconstruction challenges highlight a broader issue in the field: the scarcity of comprehensive 3D vehicle damage datasets. This scarcity necessitated our reliance on back-projected 2D metrics for evaluation. A crucial next step for validating this and similar methods involves the creation of such datasets, which would enable the use of more robust 3D-specific metrics like 3D IoU and Chamfer distance. Furthermore, expanding the evaluation to include more diverse real-world and synthetic datasets, alongside a broader comparative analysis against other 3D-GS segmentation baselines, remains an important avenue for future work to fully establish the robustness of single-view approaches.
\par
As mentioned previously, our method is similar to the rasterization algorithm from 3D-GS, given that both traverse and process a list of 2D-projected Gaussians. In future work, we could implement our algorithm using an approach similar to their rasterization. By splitting the mask region into multiple tiles, we could perform depth-buffering using multiple threads, significantly reducing the processing time.

\section{Conclusion}
\par
This work presents a complete pipeline for automatic vehicle damage detection that combines 2D and 3D segmentation in order to visualize damaged car parts on a 3D reconstruction of the vehicle. We conducted several experiments with instance segmentation models on two public damage datasets. We generated the 3D reconstruction of the vehicle using 3D Gaussian Splatting and showed the benefits of this view-synthesis method for our task. Extensive 3D segmentation experiments demonstrate the robustness of our single-view method, achieving results comparable to multi-view approaches without the associated preprocessing overhead. By eliminating the need for complex multi-view annotations, our approach paves the way for practical, scalable 3D damage analysis and other fine-grained 3D instance segmentation tasks directly from casual video captures.

\bibliographystyle{IEEEtran}
\bibliography{sample}

\begin{thebibliography}{10}
\providecommand{\url}[1]{#1}
\csname url@samestyle\endcsname
\providecommand{\newblock}{\relax}
\providecommand{\bibinfo}[2]{#2}
\providecommand{\BIBentrySTDinterwordspacing}{\spaceskip=0pt\relax}
\providecommand{\BIBentryALTinterwordstretchfactor}{4}
\providecommand{\BIBentryALTinterwordspacing}{\spaceskip=\fontdimen2\font plus
\BIBentryALTinterwordstretchfactor\fontdimen3\font minus
  \fontdimen4\font\relax}
\providecommand{\BIBforeignlanguage}[2]{{%
\expandafter\ifx\csname l@#1\endcsname\relax
\typeout{** WARNING: IEEEtran.bst: No hyphenation pattern has been}%
\typeout{** loaded for the language `#1'. Using the pattern for}%
\typeout{** the default language instead.}%
\else
\language=\csname l@#1\endcsname
\fi
#2}}
\providecommand{\BIBdecl}{\relax}
\BIBdecl

\bibitem{WangXinkuang2023}
X.~Wang, W.~Li, and Z.~Wu, ``Cardd: A new dataset for vision-based car damage
  detection,'' \emph{IEEE Transactions on Intelligent Transportation Systems},
  vol.~24, no.~7, pp. 7202--7214, 2023.

\bibitem{HuynhNhan2023}
N.~T. Huynh, N.~N. Tran, A.~T. Huynh, V.-D. Hoang, and H.~D. Nguyen, ``Vehide
  dataset: New dataset for automatic vehicle damage detection in car
  insurance,'' in \emph{2023 15th International Conference on Knowledge and
  Systems Engineering (KSE)}.\hskip 1em plus 0.5em minus 0.4em\relax IEEE,
  2023, pp. 1--6.

\bibitem{ParslovJens2024}
J.~Parslov, E.~Riise, and D.~P. Papadopoulos, ``Crashcar101: Procedural
  generation for damage assessment,'' in \emph{Proceedings of the IEEE/CVF
  Winter Conference on Applications of Computer Vision (WACV)}, January 2024.

\bibitem{KerblBernhard2023}
\BIBentryALTinterwordspacing
B.~Kerbl, G.~Kopanas, T.~Leimk{\"u}hler, and G.~Drettakis, ``3d gaussian
  splatting for real-time radiance field rendering,'' \emph{ACM Transactions on
  Graphics}, vol.~42, no.~4, July 2023. [Online]. Available:
  \url{https://repo-sam.inria.fr/fungraph/3d-gaussian-splatting/}
\BIBentrySTDinterwordspacing

\bibitem{CenJiazhong2025}
J.~Cen, J.~Fang, C.~Yang, L.~Xie, X.~Zhang, W.~Shen, and Q.~Tian, ``Segment any
  3d gaussians,'' in \emph{Proceedings of the AAAI Conference on Artificial
  Intelligence}, vol.~39, 2025, pp. 1971--1979.

\bibitem{CenJiazhong2025_2}
J.~Cen, J.~Fang, Z.~Zhou, C.~Yang, L.~Xie, X.~Zhang, W.~Shen, and Q.~Tian,
  ``Segment anything in 3d with radiance fields,'' \emph{International Journal
  of Computer Vision}, pp. 1--23, 2025.

\bibitem{ChoiSeokhun2024}
S.~Choi, H.~Song, J.~Kim, T.~Kim, and H.~Do, ``Click-gaussian: Interactive
  segmentation to any 3d gaussians,'' in \emph{European Conference on Computer
  Vision}.\hskip 1em plus 0.5em minus 0.4em\relax Springer, 2024, pp. 289--305.

\bibitem{ZhouShijie2024}
S.~Zhou, H.~Chang, S.~Jiang, Z.~Fan, Z.~Zhu, D.~Xu, P.~Chari, S.~You, Z.~Wang,
  and A.~Kadambi, ``Feature 3dgs: Supercharging 3d gaussian splatting to enable
  distilled feature fields,'' in \emph{Proceedings of the IEEE/CVF Conference
  on Computer Vision and Pattern Recognition}, 2024, pp. 21\,676--21\,685.

\bibitem{ZhuRunsong2025}
R.~Zhu, S.~Qiu, Z.~Liu, K.-H. Hui, Q.~Wu, P.-A. Heng, and C.-W. Fu,
  ``Rethinking end-to-end 2d to 3d scene segmentation in gaussian splatting,''
  in \emph{Proceedings of the Computer Vision and Pattern Recognition
  Conference}, 2025, pp. 3656--3665.

\bibitem{GuoYansong2025}
Y.~Guo, J.~Hu, Y.~Qu, and L.~Cao, ``Wildseg3d: Segment any 3d objects in the
  wild from 2d images,'' \emph{arXiv preprint arXiv:2503.08407}, 2025.

\bibitem{HuXu2025}
X.~Hu, Y.~Wang, L.~Fan, J.~Fan, J.~Peng, Z.~Lei, Q.~Li, and Z.~Zhang, ``Sagd:
  Boundary-enhanced segment anything in 3d gaussian via gaussian
  decomposition,'' \emph{arXiv preprint arXiv:2401.17857}, 2024.

\bibitem{KirillovAlex2023}
A.~Kirillov, E.~Mintun, N.~Ravi, H.~Mao, C.~Rolland, L.~Gustafson, T.~Xiao,
  S.~Whitehead, A.~C. Berg, W.-Y. Lo \emph{et~al.}, ``Segment anything,'' in
  \emph{Proceedings of the IEEE/CVF international conference on computer
  vision}, 2023, pp. 4015--4026.

\bibitem{RedmonJoseph2016}
J.~Redmon, S.~Divvala, R.~Girshick, and A.~Farhadi, ``You only look once:
  Unified, real-time object detection,'' in \emph{Proceedings of the IEEE
  conference on computer vision and pattern recognition}, 2016, pp. 779--788.

\bibitem{SchonbergerJohannes2016}
J.~L. Sch\"{o}nberger and J.-M. Frahm, ``Structure-from-motion revisited,'' in
  \emph{Conference on Computer Vision and Pattern Recognition (CVPR)}, 2016.

\bibitem{perez2024automated}
S.~A. P{\'e}rez-Zarate, D.~Corzo-Garc{\'\i}a, J.~L. Pro-Mart{\'\i}n, J.~A.
  {\'A}lvarez-Garc{\'\i}a, M.~A. Mart{\'\i}nez-del Amor, and
  D.~Fern{\'a}ndez-Cabrera, ``Automated car damage assessment using computer
  vision: Insurance company use case,'' \emph{Applied Sciences}, vol.~14,
  no.~20, p. 9560, 2024.

\bibitem{DwivediMahavir2021}
M.~Dwivedi, H.~S. Malik, S.~Omkar, E.~B. Monis, B.~Khanna, S.~R. Samal,
  A.~Tiwari, and A.~Rathi, ``Deep learning-based car damage classification and
  detection,'' in \emph{Advances in artificial intelligence and data
  engineering: Select proceedings of AIDE 2019}.\hskip 1em plus 0.5em minus
  0.4em\relax Springer, 2021, pp. 207--221.

\bibitem{ChenQiqiang2020}
Q.~Chen, X.~Gan, W.~Huang, J.~Feng, and H.~Shim, ``Road damage detection and
  classification using mask r-cnn with densenet backbone,'' \emph{Computers,
  Materials, Continua}, vol.~65, pp. 2201--2215, 01 2020.

\bibitem{HeKaiming2017}
K.~He, G.~Gkioxari, P.~Doll{\'a}r, and R.~Girshick, ``Mask r-cnn,'' in
  \emph{Proceedings of the IEEE international conference on computer vision},
  2017, pp. 2961--2969.

\bibitem{ChenHanxiao}
H.~Chen, ``Car damage detection and patch-to-patch self-supervised image
  alignment,'' \emph{arXiv preprint arXiv:2403.06674}, 2024.

\bibitem{LeeDonggeun2024}
\BIBentryALTinterwordspacing
D.~Lee, J.~Lee, and E.~Park, ``Automated vehicle damage classification using
  the three-quarter view car damage dataset and deep learning approaches,''
  \emph{Heliyon}, vol.~10, no.~14, p. e34016, 2024. [Online]. Available:
  \url{https://www.sciencedirect.com/science/article/pii/S2405844024100473}
\BIBentrySTDinterwordspacing

\bibitem{Perez-ZarateSergio2024}
\BIBentryALTinterwordspacing
S.~A. Pérez-Zarate, D.~Corzo-García, J.~L. Pro-Martín, J.~A. Álvarez
  García, M.~A. Martínez-del Amor, and D.~Fernández-Cabrera, ``Automated car
  damage assessment using computer vision: Insurance company use case,''
  \emph{Applied Sciences}, vol.~14, no.~20, 2024. [Online]. Available:
  \url{https://www.mdpi.com/2076-3417/14/20/9560}
\BIBentrySTDinterwordspacing

\bibitem{RamazhanMuhammad2025}
\BIBentryALTinterwordspacing
M.~R.~S. Ramazhan, A.~Bustamam, and R.~A. Buyung, ``Smart car damage assessment
  using enhanced yolo algorithm and image processing techniques,''
  \emph{Information}, vol.~16, no.~3, 2025. [Online]. Available:
  \url{https://www.mdpi.com/2078-2489/16/3/211}
\BIBentrySTDinterwordspacing

\bibitem{YogeshwaranS2025}
Y.~S, R.~J. J, and R.~Vasanthi, ``Yolov8-powered real-time car damage
  detection,'' in \emph{2025 3rd International Conference on Intelligent Data
  Communication Technologies and Internet of Things (IDCIoT)}, 2025, pp.
  2223--2228.

\bibitem{DuXiaobiao2024}
X.~Du, H.~Sun, M.~Lu, T.~Zhu, and X.~Yu, ``Dreamcar: Leveraging car-specific
  prior for in-the-wild 3d car reconstruction,'' \emph{IEEE Robotics and
  Automation Letters}, 2024.

\bibitem{LiCongcong2025}
C.~Li, J.~Wang, X.~Wang, X.~Zhou, W.~Wu, Y.~Zhang, and T.~Cao, ``Car-gs:
  Addressing reflective and transparent surface challenges in 3d car
  reconstruction,'' \emph{arXiv preprint arXiv:2501.11020}, 2025.

\bibitem{AuclairAdrien2007}
A.~Auclair, L.~Cohen, and N.~Vincent, ``A robust approach for 3d cars
  reconstruction,'' vol. 4522, 06 2007, pp. 183--192.

\bibitem{WangBo2024}
B.~Wang, Q.~Wu, H.~Wang, L.~Hu, and B.~Li, ``3d surface reconstruction of car
  body based on any single view,'' \emph{IEEE Access}, vol.~12, pp.
  74\,903--74\,914, 2024.

\bibitem{JayawardenaSrimal}
S.~Jayawardena, ``Image based automatic vehicle damage detection,'' Ph.D.
  dissertation, Australian National University, 11 2013.

\bibitem{vanRuitenbeekR}
\BIBentryALTinterwordspacing
R.~E. van Ruitenbeek and S.~Bhulai, ``Multi-view damage inspection using
  single-view damage projection,'' \emph{Machine Vision and Applications},
  vol.~33, no.~3, p.~46, 2022. [Online]. Available:
  \url{https://doi.org/10.1007/s00138-022-01295-w}
\BIBentrySTDinterwordspacing

\bibitem{MildenhallBen2021}
B.~Mildenhall, P.~P. Srinivasan, M.~Tancik, J.~T. Barron, R.~Ramamoorthi, and
  R.~Ng, ``Nerf: Representing scenes as neural radiance fields for view
  synthesis,'' \emph{Communications of the ACM}, vol.~65, no.~1, pp. 99--106,
  2021.

\bibitem{JainUmangi2024}
U.~Jain, A.~Mirzaei, and I.~Gilitschenski, ``Gaussiancut: Interactive
  segmentation via graph cut for 3d gaussian splatting,'' in \emph{The
  Thirty-eighth Annual Conference on Neural Information Processing Systems},
  2024.

\bibitem{ShenQiuhong2024}
Q.~Shen, X.~Yang, and X.~Wang, ``Flashsplat: 2d to 3d gaussian splatting
  segmentation solved optimally,'' in \emph{European Conference on Computer
  Vision}.\hskip 1em plus 0.5em minus 0.4em\relax Springer, 2024, pp. 456--472.

\bibitem{JocherGlenn2023}
G.~Jocher, J.~Qiu, and A.~Chaurasia, ``Ultralytics yolo,'' 2023, page:
  \url{https://github.com/ultralytics/ultralytics}, Version: 8.0.0, License:
  AGPL-3.0.

\bibitem{SchonbergerJohannes2016_2}
J.~L. Sch\"{o}nberger, E.~Zheng, M.~Pollefeys, and J.-M. Frahm, ``Pixelwise
  view selection for unstructured multi-view stereo,'' in \emph{European
  Conference on Computer Vision (ECCV)}, 2016.

\bibitem{DaiJifeng2017}
J.~Dai, H.~Qi, Y.~Xiong, Y.~Li, G.~Zhang, H.~Hu, and Y.~Wei, ``Deformable
  convolutional networks,'' in \emph{Proceedings of the IEEE international
  conference on computer vision}, 2017, pp. 764--773.

\bibitem{MirazaeiAshkan2023}
A.~Mirzaei, T.~Aumentado-Armstrong, K.~G. Derpanis, J.~Kelly, M.~A. Brubaker,
  I.~Gilitschenski, and A.~Levinshtein, ``{SPIn-NeRF}: Multiview segmentation
  and perceptual inpainting with neural radiance fields,'' in \emph{CVPR},
  2023.

\bibitem{ToffaninPiero2024}
P.~Toffanin, ``Opensplat,'' 2024, page:
  \url{https://github.com/pierotofy/OpenSplat}, Version: 1.1.4, License:
  AGPL-3.0.

\bibitem{KnapitschArno2017}
A.~Knapitsch, J.~Park, Q.-Y. Zhou, and V.~Koltun, ``Tanks and temples:
  Benchmarking large-scale scene reconstruction,'' \emph{ACM Transactions on
  Graphics}, vol.~36, no.~4, 2017.

\end{thebibliography}
\end{document}